\documentclass[journal,twoside,web]{ieeecolor}
\usepackage{generic}
\usepackage{cite}
\usepackage{amsmath,amssymb,amsfonts}
\usepackage{algorithmic}
\usepackage{graphicx}
\usepackage{algorithm,algorithmic}
\usepackage{hyperref}
\hypersetup{hidelinks=true}
\usepackage{textcomp}
\def\BibTeX{{\rm B\kern-.05em{\sc i\kern-.025em b}\kern-.08em
    T\kern-.1667em\lower.7ex\hbox{E}\kern-.125emX}}
\begin{document}

\title{Towards More Efficient Depression Risk Recognition via Gait}

\author{Min Ren, Muchan Tao, Xuecai Hu, Xiaotong Liu, Qiong Li, Yongzhen Huang$^*$
\thanks{M. Ren, M. Tao, X. Hu, X. Liu, Q. Li, and Y. Huang are with the School of Artificial Intelligence, Beijing Normal University, Beijing 100875, China (e-mail: renmin@bnu.edu.cn, mooochann@163.com, huxuecai@bnu.edu.cn, 202131081019@mail.bnu.edu.cn, 202131081018@mail.bnu.edu.cn, huangyongzhen@bnu.edu.cn).}
%\thanks{Q. Ji is with WATRIX.AI (e-mail: qing.ji@watrix.ai).}
%\thanks{$\dag$ Equal Contribution}
\thanks{$*$ Corresponding author: Yongzhen Huang.}}

%\author{First A. Author, \IEEEmembership{Fellow, IEEE}, Second B. Author, and Third C. Author Jr., \IEEEmembership{Member, IEEE}
%\thanks{This paragraph of the first footnote will contain the date on 
%which you submitted your paper for review. It will also contain support 
%information, including sponsor and financial support acknowledgment. For 
%example, ``This work was supported in part by the U.S. Department of 
%Commerce under Grant 123456.'' }
%\thanks{The next few paragraphs should contain 
%the authors' current affiliations, including current address and e-mail. For 
%example, First A. Author is with the National Institute of Standards and 
%Technology, Boulder, CO 80305 USA (e-mail: author@boulder.nist.gov). }
%\thanks{Second B. Author Jr. was with Rice University, Houston, TX 77005 USA. He is 
%now with the Department of Physics, Colorado State University, Fort Collins, 
%CO 80523 USA (e-mail: author@lamar.colostate.edu).}
%\thanks{Third C. Author is with 
%the Electrical Engineering Department, University of Colorado, Boulder, CO 
%80309 USA, on leave from the National Research Institute for Metals, 
%Tsukuba, Japan (e-mail: author@nrim.go.jp).}}

\maketitle

\begin{abstract}

% 抑郁症是危害最为广泛的心理疾病，全球有超过280 million人受到抑郁症的影响。
% Early detection and timely intervention is crucial for promoting remission, preventing relapse, and mitigating the emotional and financial burdens。
% 然而，抑郁症患者in the primary care setting常常得不到诊断和正确的治疗。
% 这是由于抑郁症 unlike many physiological illnesses,  缺乏objective indicators to recognize depression risk，而且现有的抑郁风险识别方法时间和人力成本很高。
% 步态已经被证明与抑郁风险之间存在关联关系，能够作为一种Objective biomarker，并且具有数据采集高效方便的优势。
% 但是现有的基于步态识别抑郁风险的方法都只在小规模私有数据集上做了验证，缺乏供研究用的大规模公开数据集。
% 同时，这些方法基本都局限于hand-crafted方法，human gait is a complex form of motion, and hand-crafted gait features often only capturing a fraction of the intricate associations between gait and depression risk。
% 因此，本文首先构建了一个包含1200多人、40000多个步态序列，涵盖6个视角、3种着装的大规模步态数据库，并提供了两种常用心理量表的得分作为抑郁风险标注。
% 而后，本文提出了一个基于深度学习的 depression risk recognition model，克服了hand-crafted方法的不足。
% 通过在所构建的大规模数据库上的实验验证了做提出的方法的有效性，并且给出了很多instructive insights are given in our paper, which indicates the significant potential of gait-based depression risk recognition.
Depression, a highly prevalent mental illness, affects over 280 million individuals worldwide.
Early detection and timely intervention are crucial for promoting remission, preventing relapse, and alleviating the emotional and financial burdens associated with depression.
However, patients with depression often go undiagnosed in the primary care setting.
Unlike many physiological illnesses, depression lacks objective indicators for recognizing depression risk, and existing methods for depression risk recognition are time-consuming and often encounter a shortage of trained medical professionals.
The correlation between gait and depression risk has been empirically established.
Gait can serve as a promising objective biomarker, offering the advantage of efficient and convenient data collection.
However, current methods for recognizing depression risk based on gait have only been validated on small, private datasets, lacking large-scale publicly available datasets for research purposes.
Additionally, these methods are primarily limited to hand-crafted approaches.
Gait is a complex form of motion, and hand-crafted gait features often only capture a fraction of the intricate associations between gait and depression risk.
Therefore, this study first constructs a large-scale gait database, encompassing over 1,200 individuals, 40,000 gait sequences, and covering six perspectives and three types of attire.
Two commonly used psychological scales are provided as depression risk annotations.
Subsequently, a deep learning-based depression risk recognition model is proposed, overcoming the limitations of hand-crafted approaches. 
Through experiments conducted on the constructed large-scale database, the effectiveness of the proposed method is validated, and numerous instructive insights are presented in the paper, highlighting the significant potential of gait-based depression risk recognition.

\end{abstract}

\begin{IEEEkeywords}

Depression Risk, Gait Analysis, Deep Learning, Dynamic Features
%Enter key words or phrases in alphabetical order, separated by commas. Using the IEEE Thesaurus can help you find the best standardized keywords to fit your article. Use the thesaurus access request form for free access to the IEEE Thesaurus: \underline{https://www.ieee.org/publications/services/thesaurus-acce}\\
%\underline{ss-page.com.}
\end{IEEEkeywords}

%%%%%%%%%%%%%%%%%%%%%%%%%%%%%%%%%%%%%%%%%%%%%%%%%%%%%%%%%%%%
%%%%%%%%%%%%%%         Introduction       %%%%%%%%%%%%%%%%%%
%%%%%%%%%%%%%%%%%%%%%%%%%%%%%%%%%%%%%%%%%%%%%%%%%%%%%%%%%%%%

\section{Introduction}

Depression is recognized by the World Health Organization (WHO) as one of the primary contributors to the global disease burden\cite{WHO}.
Within the clinical framework, the American Psychiatric Association's Diagnostic Statistical Manual of Mental Disorders- Fifth Edition (DSM-5) classifies depression as persistent depressive mood or loss of interest and pleasure in activities.
It affects approximately 280 million individuals worldwide, comprising approximately 3.8\% of the global population\cite{vizhub}.
Depression profoundly impacts individuals across multiple domains, including a decline in their overall quality of life\cite{kennedy2001quality, gaynes2002depression}, compromised social functioning\cite{hirschfeld2000social, hansson2002quality}, detrimental effects on physical health\cite{bourassa2017social,roshanaei2009longitudinal}, and an elevated susceptibility to suicide\cite{hawton2013risk,center2003confronting}.

Recognizing the significance of early detection and timely intervention is crucial for promoting remission, preventing relapse, and mitigating the emotional and financial burdens associated with this condition\cite{coulehan1997treating,coyne1994prevalence,halfin2007depression}. 
However, depression is notably underdiagnosed\cite{coulehan1997treating,weich2007attitudes,souery2007clinical}.
A survey involving 33,653 physician-patient interactions reveal that in the primary care setting, less than 5\% of adults are screened for depression\cite{akincigil2017national}.
And it has been reported that at least 25\% of patients go undiagnosed\cite{barbui2006identification}, with a significant majority of those seeking assistance from primary care physicians not receiving suitable treatment, particularly in low-income and middle-income countries\cite{young2001quality,kessler2005prevalence}.

The main challenge in screening depression in the primary care is the absence of objective indicators to recognize depression risk\cite{belmaker2008major}.
Unlike many physiological illnesses, depression lacks precise biomarkers, clinicians primarily rely on clinical criteria such as psychological questionnaires and patients’ self-reports\cite{greden2001burden}. 
This introduces potential issues:
Individuals may provide biassed or inaccurate information due to personal inclinations, social expectations or difficulties in recalling past experiences\cite{mohr2006barriers, kravitz2011relational, cassano2002depression}.
Approximately 50\% of patients have been observed to negate experiencing depressive feelings\cite{belmaker2008major}.
Besides, variability in linguistic expression can hinder accurate communication of emotions or experiences\cite{bell2011suffering, johnstone2001stigma, moyle2002unstructured, delbaere2010determinants}.
These issues can lead to incorrect depression risk recognition, and missing the optimal window for further treatment\cite{rost2004effect, docherty1997barriers, dowrick2013medicalising}.

Another primary issue with questionnaire-based approaches for depression risk recognition pertains to inefficiency.
Psychological assessments are time-consuming and often encounter a shortage of trained medical professionals, resulting in prolonged referral processes.
Consequently, this may hinder timely psychological intervention, potentially exacerbating mental health issues\cite{cheng2021addressing, butryn2017shortage}.
In addition, psychological questionnaires are typically intended for periodic, rather than frequent, evaluations, often yielding only one or two measurements annually.
This limited frequency may not effectively capture long-term condition trends in a timely manner, in light of the high relapse rates associated with depression: 21\% at 12 months, 30\% at 2 years, and 42\% at 5 years\cite{kanai2003time}.

Gait has been shown to be an essential manifestation of depression risk\cite{sobin1997psychomotor, schrijvers2008psychomotor, fried2014impact}.
In particular, gait is modulated by the advanced neural center\cite{takakusaki2017functional}, which is also implicated with the pathophysiology of depression\cite{walther2019utility}.
Many researches have demonstrate the association between depression risk and gait characteristics.
Notably, specific abnormalities in gait, such as decreased vertical head movement\cite{michalak2009embodiment}, reduced range of motion in limbs, and a decelerated pace\cite{lemke2000spatiotemporal}, are stable indications of depression.
Hence, the gait-based depression risk recognition offers advantages compared to traditional psychological assessment methods:
\begin{itemize}
    \item \textbf{Objective biomarker}: 
    Gait serves as an objective biomarker for recognizing the risk of depression within the primary care setting, thereby mitigating the influence of subjective biases, societal expectations, or challenges in recollecting past experiences, which often lead to inaccurate information.

    \item \textbf{High efficiency}:
    Gathering gait data through cameras proves to be a highly efficient method, surpassing the time-consuming process of administering psychological questionnaires.
    This approach eliminates the requirement for trained professionals to collect data, thereby reducing both time and manpower costs.
    
\end{itemize}

%%

% 虽然研究者们已经对步态与抑郁风险之间的关系进行了初步的研究，也有研究者试图基于步态特征对抑郁风险进行识别。
% 但是已有的研究存在两个重要的缺陷：
% 首先，这些研究所基于的数据量都较少，而且较为单一，通常并不包含不同视角、不同着装的步态数据。
% 这使得这些研究成果的适用性有待证实，即这些研究所得出的结论并不一定适用于更加一般的步态数据。
% 而且这些数据集几乎都是不能公开获得的，这不仅是的其研究难以复现，而且阻碍了这一领域的发展。
% 第二，这些研究通常采用手工设计的步态特征进行分析。
% 然而人体的行走是一种复杂的运动形式，手工设计的步态特征往往只能简单地表达步态的一个方面，很难全面地反应步态与抑郁风险的关联。
Despite preliminary investigations into the relationship between gait and depression risk, as well as attempts to recognize depression risk based on gait features, existing research suffers from two significant limitations.
Firstly, these studies are often constrained by limited and homogeneous gait data, typically lacking the inclusion of gait data captured from different perspectives and with varying attire.
Consequently, the generalizability of these research findings remains to be established, as the conclusions drawn may not necessarily apply to more diverse gait data.
Furthermore, the accessibility of these datasets is severely restricted, impeding research reproducibility and hindering progress in the field.
This limitation not only undermines the ability to validate previous research but also hampers advancements in this domain.
Secondly, these studies commonly rely on hand-crafted gait features for analysis.
However, human gait is a complex form of motion, and hand-crafted gait features often provide a limited representation, only capturing a fraction of the intricate associations between gait and depression risk.

%%

% 为此，我们首先收集了一个目前为止最大的给予步态的抑郁风险数据集，其中包含1200多名受试者，4万多个步态序列，并且包含了多种视角、多种着装。
% 并采用两种初诊中最常用的抑郁风险评估量表作为被试者的抑郁风险标注。
% 在这一数据集的基础上，我们在本文中提出了一个新颖的数据驱动的基于深度学习的抑郁风险识别方法。
% 这一方法能够克服手工设计的特征带来的问题，自主学习抑郁风险相关的步态特征建模方法，为步态与抑郁风险关联关系的研究提供了全新的研究思路。
To address these limitations, we have established the largest gait-based dataset for depression risk recognition to date, encompassing over 1,200 subjects and more than 40,000 gait sequences.
This dataset incorporates diverse perspectives and attire variations, providing a comprehensive representation of gait data.
The depression risk of the participants was annotated using two commonly used depression risk assessment scales in the primary care.

Leveraging this rich dataset, we propose a novel data-driven approach for depression risk recognition based on deep learning in this paper.
The proposed method overcomes the problems associated with hand-crafted features by autonomously learning gait features that are relevant to depression risk.
There are two notable characteristics in depression risk related gait features: 
Firstly, the main features associated with depression risk are dynamic features, which refer to the temporal aspects of gait during the walking process.
Secondly, gait features that are associated with depression risk can manifest as either local details or involve the entire body.
Based on these two characteristics, this paper proposes a deep learning model with dynamic feature modeling as its core.
In the feature extraction process, both local dynamic features and global dynamic features are effectively integrated.
This innovative methodology offers a fresh perspective for investigating the association between gait and depression risk.

%%

% 贡献：
% 1. To the best of our knowledge, we built the first large-scale gait-based depression risk recognition dataset.
% 这一数据集将最为一个benchmark，能够推动抑郁风险识别领域的发展，最终有助于提高抑郁患者的初诊筛查效率，让更多的患者能够获得帮助。
% 2. 本文提出了一个深度学习抑郁风险识别方法。
% 基于对抑郁相关的步态特征的洞见，我们提出了一个新的动态特征建模结构，并将局部特征与全局特征进行了有效融合。
% 3. Abundant experiments further explore the abnormal gait patterns correlating to depression risk, which provides instructive insights for the research on depression.

The contributions of this paper can be summarized as follows:
\begin{itemize}
    \item We build a large-scale dataset for gait-based depression risk recognition.
    This dataset serves as a benchmark, propelling advancements in the field of depression risk recognition.
    Ultimately, it aims to enhance the efficiency of depression screening in the primary care, ensuring that more individuals receive the assistance they need.

    \item Based on the insights on depression-related gait features, we introduce a deep learning model with dynamic feature modeling at its essence.
    This model adeptly merges local and global features, yielding a comprehensive integration.

    \item A plethora of experiments are conducted to delve deeper into the aberrant gait patterns that are associated with the risk of depression.
    These experiments offer enlightening perspectives that contribute to the field of depression research.
\end{itemize}

The remainder of this paper is organized as follows:
Section~\ref{sec:RelateWork} presents a brief literature review of the related work.
Section~\ref{sec:Dataset} introduce the gait-based dataset for depression risk recognition.
The proposed recognition model are described in detail in Section~\ref{sec:Method}.
The configurations and results of experiments are presented in Section~\ref{sec:Experiments}.
Finally, the conclusion of this paper is summarized in Section~\ref{sec:Conclusion}.

%

%

%%%%%%%%%%%%%%%%%%%%%%%%%%%%%%%%%%%%%%%%%%%%%%%%%%%%%%%%%%%%
%%%%%%%%%%%%%%         Related Work       %%%%%%%%%%%%%%%%%%
%%%%%%%%%%%%%%%%%%%%%%%%%%%%%%%%%%%%%%%%%%%%%%%%%%%%%%%%%%%%

\section{Related Work}
\label{sec:RelateWork}

% --------------- Machine Learning based Depression Risk Recognition ------------- %
\subsection{Machine Learning based Depression Risk Recognition}

% 由于抑郁症带来的巨大危害，研究者们在抑郁风险识别领域作出了许多研究。
% 机器学习方法的应用推动了这一领域的发展。
% 基于机器学习的抑郁风险识别方法涵盖了多种模态的数据。
Due to the significant harm caused by depression, researchers have conducted numerous studies in the field of depression risk recognition.
The application of machine learning methods has propelled advancements in this field.
Machine learning-based approaches for recognizing depression risk encompass various modalities of data.

%%

% 脑电信号、心电信号等生理信号可被用于抑郁症的风险识别和辅助诊断 [5-9]。
% 抑郁症会带来异常的大脑活动，这些异常活动可以通过脑电信号反映出来。
% 同时，由于相关的自主神经功能发生障碍，抑郁症发作可能伴随恶心、 呕吐、胸部紧张和出汗等症状[10]，这些变化可以反映在心电信号上。
% 主成分分析、支持向量机、随机森林以及LSTM等常用的机器学习方法都被用于分析这些生理信号与抑郁风险之间的关联关系。
Physiological signals such as electroencephalogram (EEG) and electrocardiogram (ECG) can be utilized for the recognition of depression risk.
Depression is associated with abnormal brain activity, which can be reflected through EEG signals.
Additionally, due to disruptions in the autonomic nervous system, depressive episodes may be accompanied by symptoms such as nausea, vomiting, chest tightness, and sweating\cite{nilsonne2021eeg}, which can be observed in ECG signals.
Common machine learning methods, including principal component analysis (PCA), support vector machines (SVM), random forests, and long short-term memory (LSTM), have been employed to analyze the relationship between these physiological signals and depression risk\cite{schnyer2017evaluating, ramasubbu2016accuracy, vai2020predicting}.

%% 

% 脑部成像技术也是抑郁风险识别的手段之一。
% 主要包括核磁共振 (MRI)、功能性近红外光谱成像(fNIRS)以及为正电子发射断层扫描(PET) 等。
% 抑郁症可能会导致大脑的结构和功能异常，例如海马体萎缩，神经递质发 生变化，并引起慢性炎症等。
% 脑部成像技术能够通过对这些症状进行的检测来评估患者的病情。
% 为了实现自动化地处理和分析脑部图像，机器学习领域中的图像处理方法被引入了这一领域[11-15]。
Brain imaging techniques are also among the means used for recognizing depression risk.
This primarily includes magnetic resonance imaging (MRI), functional near-infrared spectroscopy (fNIRS), and positron emission tomography (PET), among others.
Depression can potentially lead to structural and functional abnormalities in the brain, such as hippocampal atrophy, neurotransmitter changes, and chronic inflammation.
Brain imaging techniques can assess the condition of patients by detecting these symptoms.
In order to achieve automated processing and analysis of brain images, image processing methods from the field of machine learning have been introduced in this domain\cite{shim2019machine, jiang2016predictability, sato2015machine, wei2021functional}.

%%

% 此外，抑郁症可能会导致患者的面部表情模式、说话声音发生改变。
% 有研究者提出将表情分析技术[17]和语音分析技术[16]用于抑郁风险识别。
Furthermore, depression can potentially lead to changes in facial expression patterns and speech patterns of patients.
Researchers have proposed the use of facial expression analysis techniques\cite{zhou2018visually} and speech analysis techniques\cite{ma2016depaudionet} for the recognition of depression risk.

%%

% 上述的基于机器学习方法探索了抑郁风险识别的各种途径。
% 然而这些方法所采用的数据模态均需要受试者在特定场景、采用特定的设备进行数据采集，需要受试者高度配合，导致这些方法并不比传统的psychological questionnaires更加高效。
The aforementioned exploration of depression risk recognition through machine learning methods has explored various approaches.
However, the data modalities used by these methods require participants to undergo data collection in specific scenarios and with specific devices, demanding a high level of cooperation from the participants. 
As a result, these methods incur higher time and manpower costs, ultimately making them no more efficient than traditional psychological questionnaires in practice.

% --------------- Gait Analysis ------------- %
\subsection{Gait Analysis}

% 人体步态分析从研究方法可以大致分为两大类——基于模型的步态分析方法和基于表观的步态分析方法。
% 在基于模型的步态分析研究方面，比较早的代表性研究工作是 Wang et al.[28] 提出的将静态人体信息和动态人体信息进行融的方法。
% 此后，Li et al.[29]提出了一种端到端的基于模型的步态分析与识别方法，给定一个 RGB 步态序列，首先通过拟合 SMPL 模型来提取姿态和形状特征，然后将姿态和形状特征提供给深度学习模型进行分析。
% Torben et al.[34]则将图卷积网络应用于人体骨架姿态估计即步态分析。
% 基于模型的步态分析研究思路通过对人体结构进行建模，使得所提取的特征具有更好的不变性，在目标存在遮挡、噪音、尺度变换和旋转时也能取得较好的效果。
Gait analysis can be broadly categorized into two main streams of research methods: model-based gait analysis and appearance-based gait analysis.
In the realm of model-based gait analysis research, a pioneering study by Wang et al.\cite{wang2004fusion} proposed a method that combines static and dynamic human body information.
Subsequently, Li et al.\cite{li2020end} introduced an end-to-end model-based gait analysis and recognition method.
Given an RGB gait sequence, it first extracts pose and shape features by fitting the SMPL model, and then provides these features to a deep learning model for analysis.
The model-based approach in gait analysis research aims to model the structure of the human body, resulting in extracted features that possess better invariance.
These approaches demonstrate good performance even in the presence of occlusion, noise, scale variations, and rotations.

%%

% 相比之下，基于表观的步态分析方法则从人体的外观在时空中的变化规律出发，对人体步态进行分析。
% 早期的代表性研究工作包括 Kusakunnira et al.[30]针对视角变换问题提出了基于支持向量回归的视角转换模型，通过步态能量图[36]进行局部运动特征选择，通过回归来实现模型的构建。
% Bashir et al.[31] 通过高斯过程来估计每个步态序列的视角，在此基础上使用典型关联分析对不同视角下的步态序列进行建模，从而实现跨视角步态分析。
% 随着深度学习的兴起，基于深度神经网络的表观分析的方法逐渐成为了步态分析领域的主流。
% Wu et al.[33]通过深度卷积网络来学习两个输入样本之间的相似性实现步态识别，他们的研究发现深度神经网络能够有效地提取步态特征。
% 此后的步态分析工作主要将精力集中在克服表观模型带来的视角、衣着带来的步态变化上。
% Yu et al.[35]将对抗生成网络作为回归器来生成具有不变性的步态图像，以解决不同视角、衣着、携带物品情况下步态分析问题。
% Chao et al.[37]提出了 GaitSet 模型，将步态数据看作无序的步态剪影集合加以建模。
% Liao et al.[39]提出了基于姿态的步态时空分析网络来缓解衣着的影响。
% Lin et al.[44]为了更好的建模全局与局部的信息，提出了一个全局-局部特征提取器,有效的结合了全局和局部步态特征。
In contrast, appearance-based gait analysis methods analyze human gait based on the appearance of the human body.
Early representative research includes the work of Kusakunnira et al.\cite{kusakunniran2010support}, who proposed a support vector regression-based view transformation model to address the view variation problem.
Bashir et al.\cite{bashir2010cross} estimated the view of each gait sequence using Gaussian processes and used correlation analysis to model gait sequences from different view, enabling cross-view gait analysis.
With the rise of deep learning, methods based on deep neural networks for appearance analysis have gradually become mainstream in the field of gait analysis.
Wu et al.\cite{wu2016comprehensive} implemented gait recognition by learning the similarity between two input samples using deep convolutional networks.
Their research found that deep neural networks can effectively extract gait features.
Subsequent gait analysis studies have focused primarily on overcoming the challenges posed by view and attire variations.
Yu et al.\cite{yu2017gaitgan} employed generative adversarial networks as regressors to generate gait images, addressing gait analysis under different views, attire variations, and carrying objects.
Chao et al.\cite{chao2019gaitset} proposed the GaitSet model, which treats gait data as an unordered set of gait silhouettes and models them accordingly.
Song et al.\cite{song2019gaitnet} proposed a fully end-to-end gait analysis model, greatly streamlining the process of gait analysis.
Fan et al.\cite{fan2020gaitpart} proposed dividing the human body into patches to extract features separately, addressing the variations caused by different perspectives and attire through a divide-and-conquer approach. 
In response to the issue of high dimensionality in gait features, Hou et al.\cite{hou2020gait} presented a method for compressing gait features.
To better model global and local information, Lin et al.\cite{lin2021gait} proposed a global-local feature extractor.

%%

% 虽然近年来面向通用领域的步态分析方法发展迅速，但当前大部分步态分析方法主要以身份识别任务为目标，对于人体步态中所包含的与情绪、心理状态以及健康情况等细粒度的信息缺乏挖掘，尚需与相关的领域知识进行系统性融合研究。
While there has been rapid development in general gait analysis methods in recent years, most current approaches primarily focus on identity recognition tasks. 
They lack the exploration of fine-grained information contained in human gait, such as emotions, psychological states, and health conditions.
Therefore, there is a need for systematic integration of relevant domain knowledge to further research in this area.

% --------------- Data Construction ------------- %
\subsection{Gait based Depression Risk Recognition}

% 已经有一些研究者探索了步态和抑郁风险之间的关联关系。
% For example, Sloman et al. (21) analyzed photograms of a single stride during a natural walk and found that depressed patients’ walks were more slowly with a lifting motion of the leg. In contrast, healthy participants propel themselves forward with increased foot push-off. Another study used a combination of electronic walkways and photogrammetry to show that depressed patients have shorter strides and slower gait velocity than healthy controls (22). Michalak et al. (23) demonstrated that depressed individuals exhibit reduced vertical head movements, more slumped posture, and lower gait velocity than controls by using three-dimensional (3D) motion capture.
Several researchers have explored the relationship between gait and depression.
For instance, Sloman et al.\cite{sloman1982gait} conducted an analysis of photograms capturing a single stride during a natural walk and observed that individuals with depression exhibited slower walks with a lifting motion of the leg.
Another study utilized electronic walkways and photogrammetry to demonstrate that depressed patients had shorter strides and slower gait velocity compared to healthy controls\cite{lemke2000spatiotemporal}.
Michalak et al.\cite{michalak2009embodiment} utilized 3D motion capture and showed that individuals with depression displayed reduced vertical head movements, a more slumped posture, and lower gait velocity compared to controls.

%%

% 也有研究者试图通过对步态特征识别抑郁风险。
% Wang et al. employed SVM to construct a machine learning model to recognitze depression risk through spatio- temporal, time-domain, and frequency-domain features.
% Lu et al. 提出了一种基于人体skeleton的模型，试图通过 Fast Fourier Transforms (FFT)实现抑郁风险相关的步态特征提取。
% Fang et al 则采用随机森林试图识别抑郁风险。
Some researchers have also attempted to recognize depression risk through the recognition of gait features.
For example, Wang et al.\cite{wang2020gait} utilized Support Vector Machines (SVM) to construct a machine learning model that could recognize depression risk by extracting spatio-temporal, time-domain, and frequency-domain features.
Lu et al.\cite{lu2021new} proposed a model based on human body skeleton, aiming to extract gait features related to depression risk through Fast Fourier Transforms (FFT).
Fang et al.\cite{fang2019depression}, on the other hand, employed Random Forests in an attempt to recognize depression risk.

%%

% 但是已有的研究存在两个重要的缺陷：
% Firstly, these studies are often constrained by limited and homogeneous gait data, typically lacking the inclusion of gait data captured from different perspectives and with varying attire. Consequently, the generalizability of these research findings remains to be established, as the conclusions drawn may not necessarily apply to more diverse gait data. Furthermore, the accessibility of these datasets is severely restricted, impeding research reproducibility and hindering progress in the field. This limitation not only undermines the ability to validate previous research but also hampers advancements in this domain. Secondly, these studies commonly rely on hand-crafted gait features for analysis. However, human gait is a complex form of motion, and hand-crafted gait features often provide a limited representation, only capturing a fraction of the intricate associations between gait and depression risk.
However, the existing studies suffer from two noteworthy limitations:
Firstly, these studies are often constrained by a dearth of heterogeneous gait data, which frequently fails to incorporate varying perspectives and attire.
Consequently, the generalizability of their findings remains uncertain, as the conclusions drawn may not be readily applicable to a broader range of gait data.
Secondly, these studies commonly rely on hand-crafted gait features for analysis.
Nevertheless, human gait is an intricate and multifaceted form of motion, and such manually constructed features often offer a limited portrayal, capturing only a fraction of the intricate correlations between gait and the risk of depression.

%%%%%%%%%%%%%%%%%%%%%%%%%%%%%%%%%%%%%%%%%%%%%%%%%%%%%%%%%%%%
%%%%%%%%%%%%%%         Dataset       %%%%%%%%%%%%%%%%%%
%%%%%%%%%%%%%%%%%%%%%%%%%%%%%%%%%%%%%%%%%%%%%%%%%%%%%%%%%%%%

\section{Dataset}
\label{sec:Dataset}

% 为了探索基于步态的抑郁风险识别方法，我们构建了一个全新的数据集。本节对数据集的构建、数据预处理以及数据集的统计信息进行介绍。
%
To embark upon the exploration of gait-based depression risk recognition method, we establish a novel dataset.
In this section, we present the construction of the dataset, the preprocessing of the data, and provide an overview of the statistical information pertaining to the dataset.

%%

% --------------- Data Collection ------------- %

\subsection{Gait Collection}

\begin{figure}[t]
%\vspace{-0.1cm}
\begin{center}
\includegraphics[width=\linewidth]{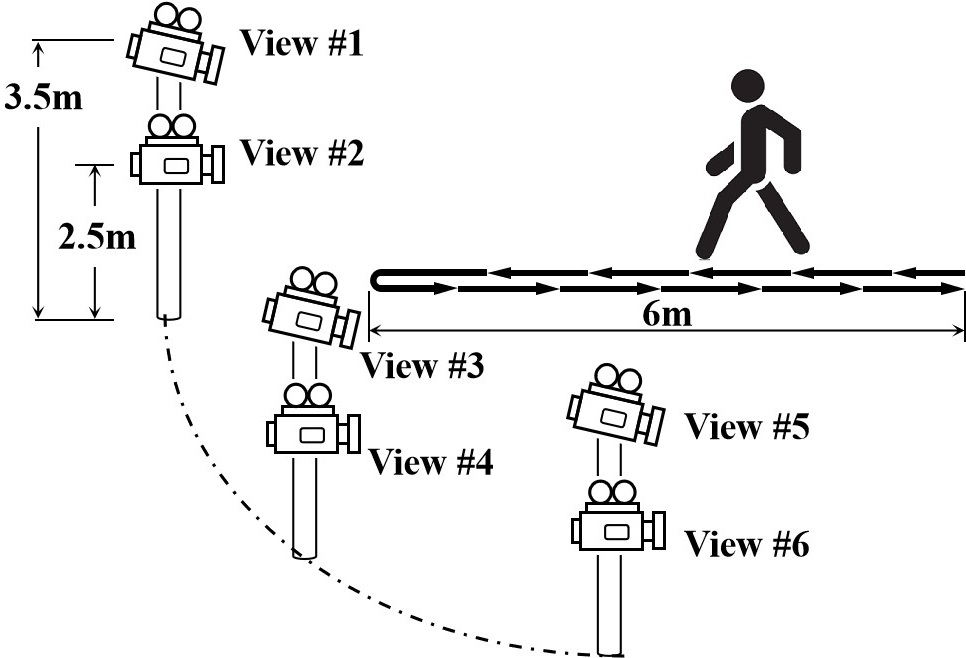}
\end{center}
   \caption{Illustration of the gait data collection process. The participants are directed to walk along a predefined path while being captured by six cameras, each offering a unique view.}
\label{fig:gait_collect}
%\vspace{-0.6cm}
\end{figure}

% 步态数据的采集对象是成年的志愿者，这些志愿者对于我们的研究内容是知情同意的。
% 志愿者在采集过程中被要求按照指定的轨迹行走，我们采用6个摄像机，从6个不同是视角对其行走过程进行拍摄。
% 六个摄像机被分为三组，每组包含一高一低两个摄像机，分别从距离地面2.5米和3.5米的高度进行拍摄，以便在垂直方向上得到不同视角的数据。
% 这三组摄像机被均匀分布在正面到侧面的圆弧上，以捕捉不同的水平视角的步态视频。
% 指定的行走轨迹是长度为6米的线段，在一次采集中，志愿者被要求从线段的这一端走到另一端，而后在返回到起始点。
% 每位志愿者被要求进行三次采集，每次采集都穿着不同的着装，包括穿着外套、不穿着外套以及背包三种情况，以便获得不同着装条件下的步态视频。
% 
% 画一个摄像机分布示意图
The subjects chosen for the collection of gait data are consenting adult volunteers who have been informed about the nature of our research endeavors.
These volunteers are instructed to traverse a predetermined walking route while being recorded by six cameras placed strategically from distinct views, as shown in Fig.~\ref{fig:gait_collect}.
The six cameras are divided into three groups, each consisting of one high and one low camera.
The two cameras of each group are positioned at heights of 2.5 meters and 3.5 meters above the ground, respectively, in order to capture data from different vertical views.
These three groups of cameras are evenly distributed along an arc ranging from the front to the side, enabling the capture of gait videos from various horizontal views.
The designated walking route spans a length of six meters, requiring the participants to traverse from one end to the other and subsequently return to the starting point.
Each volunteer is required to undergo three data collection sessions, with each session involving a distinct attire.
%
% 通过上述采集过程，便得到了六种视角、三种着装的视频步态数据
Each volunteer is required to undergo three sessions of data collection, with each session involving different attire, including wearing a coat, not wearing a coat, and carrying a backpack.
This is done to obtain gait videos under varied dressing conditions.

%%

\begin{figure}[t]
%\vspace{-0.1cm}
\begin{center}
\includegraphics[width=\linewidth]{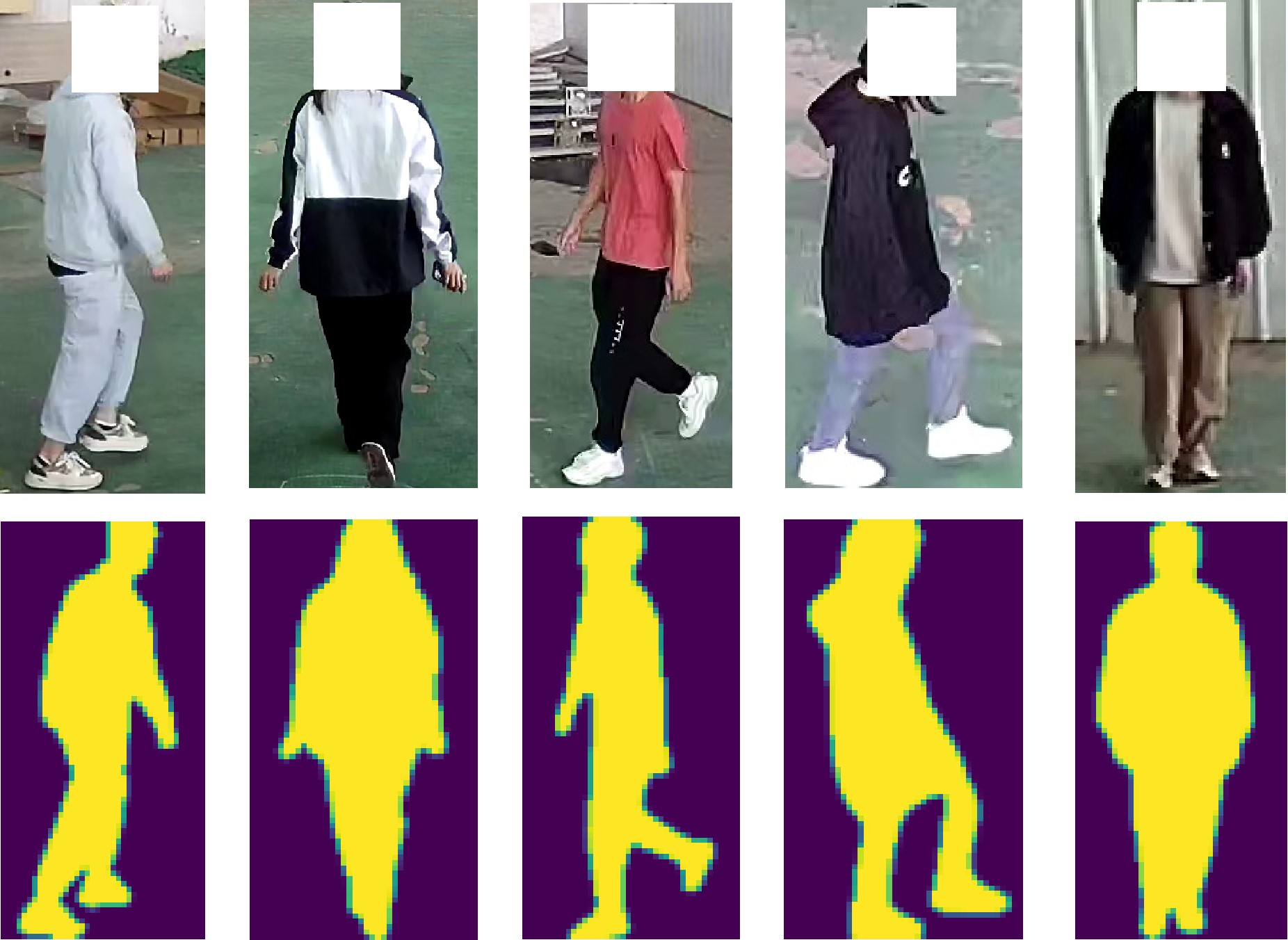}
\end{center}
   \caption{Examples of the collected gait data. Through pre-processing, only the silhouette data of the human body is retained, ensuring the effective protection of the subjects' privacy information.}
\label{fig:data_example}
%\vspace{-0.6cm}
\end{figure}

% 对视频步态数据的预处理主要包含两个步骤：
% 首先在视频中对采集对象进行跟踪
% 而后，为了尽可能保护被采集者的隐私，我们对跟踪得到的tracklet进行人体分割，并只保留分割得到的剪影，其中仅仅包含人体的轮廓信息。
% 在完成预处理之后，原始的视频数据被不可恢复地删除。
% 通过上述过程被采集对象的隐私得到了很好的保护。
% 
% 画一个预处理步骤图
The pre-processing of the video gait data primarily involves two steps:
Firstly, tracking the subjects in the videos.
Subsequently, in order to preserve the privacy of the individuals being captured, we perform human segmentation on the tracked tracklets, retaining only the silhouettes that encompass the contour information of the human body, as shown in Fig.~\ref{fig:data_example}.
%
%Once the pre-processing is completed, the original video data is irreversibly deleted.
%
\textbf{The private information is eliminated during data pre-processing, and the original data is permanently removed.
Through the aforementioned process, the privacy of the subjects involved has been effectively safeguarded}.

% 可以增加的内容：关于数据清洗的说明

% --------------- Depression Risk Annotation ------------- %

\subsection{Depression Risk Annotation}

% 两个广泛使用的量表被用于对受试者的抑郁风险进行标注：
% Self-Rating Depression Scale (SDS) 是最著名的抑郁自评量表之一，为美国教育卫生福利部推荐的用于精神药理学研究的量表之一，因使用简便，能直观反映病人抑郁的主观感受，当前广泛应用于抑郁症测试筛查。SDS 抑郁测试量表含有 20 个项目，通过测试可以快速了解受试者是否存在抑郁风险。
% 9-question Patient Health Questionnairer (PHQ-9) 是一种在门诊广泛使用的量表，用于对受试者的抑郁风险进行快速初步筛查。
% 在进行抑郁风险识别实验时，受试者根据量表得分被分为有抑郁风险的实验组以及没有抑郁风险的对照组。
% 
Two widely used psychological scales are used to assess the depression risk of the participants:

\begin{itemize}
    \item The Self-Rating Depression Scale (SDS): SDS is one of the most renowned self-report measures for depression.
    It is recommended by the U.S. Department of Health and Human Services for research in psychopharmacology.
    It is widely utilized for screening and assessing depression due to its ease of use and ability to reflect the subjective experience of patients.
    The SDS consists of 20 items and provides a quick assessment of an individual's risk of depression.

    \item The 9-question Patient Health Questionnaire (PHQ-9): PHQ-9 is a commonly used scale in outpatient settings for the rapid preliminary screening of depression risk in individuals.
\end{itemize}

During the depression risk identification experiment, participants are classified into an experimental group, indicating the presence of depression risk, and a control group, indicating the absence of depression risk, based on their scores on these scales.
%
% 同时满足以下两个条件的受试者被划分入实验组：SDS分数大于58，PHQ-9分数大于8。
% 同时满足以下两个条件的受试者则被划分入对照组：SDS分数小于47，PHQ-9分数小于2。
Participants who meet both of the following criteria are assigned to the experimental group: SDS score greater than 58 and PHQ-9 score greater than 8.
Participants who meet both of the following criteria are assigned to the control group: SDS score less than 47 and PHQ-9 score less than 2.

% --------------- Dataset Statistics ------------- %

\subsection{Dataset Statistics}

\begin{table}[t]
\begin{center}
\caption{The statistical information of the proposed dataset.}
\label{tab:DataStat_1}
\setlength{\tabcolsep}{5.5mm}
{
\begin{tabular}{c|c|c}
\hline
\hline
       & \bf{Depression} & \bf{Control}  \\
\hline
Gender (M:F)  & 306:300 & 431:190  \\
%\hline
Age & 18-25 & 18-25   \\
%\hline
SDS (Mean $\pm$ SD)  & 63.51 $\pm $6.87 & 34.42 $\pm$ 4.45   \\
%\hline
PHQ-9 (Mean $\pm$ SD) & 0.55 $\pm$ 0.42 & 12.75 $\pm$ 4.12   \\

\hline
\hline
\end{tabular}}
\end{center}
\end{table}

\begin{table}[t]
\begin{center}
\caption{The information of the proposed dataset and comparison to existing datasets.}
\label{tab:DataStat_2}
\setlength{\tabcolsep}{2.3mm}
{
\begin{tabular}{c|c|c|c|c|c}
\hline
\hline
       & \bf{Depression} & \bf{Control} & \bf{View} & \bf{Attire} & \bf{Sequence} \\
\hline
Fang\cite{fang2019depression}  & 43 & 52 & 1 & 1 & -  \\

 Yuan\cite{yuan2019depression}  & 54 & 47 & 1 & 1 & -  \\

  Wang\cite{wang2020gait}  & 126 & 121 & 1 & 1 & -  \\

  Lu\cite{lu2021new}  & 86 & 114 & 4 & 1 & -  \\

\hline
 \bf{Ours} & \bf{606} & \bf{621} & \bf{6} & \bf{3} & \bf{40281}  \\

\hline
\hline
\end{tabular}}
\end{center}
\end{table}

% 本文所提出的数据集的统计信息如表所示。
% 其与现有的数据集的情况的对比如表所示。
% 在现有的数据集中，本文所提的数据集受试者数量最多，步态序列数也最多，同时包含最多的视角以及着装情况。
% 由于现有的数据集大多涉及到受试者的隐私，因此并不能公开获得，这严重阻碍了这一领域的发展。
% 为此，我们在数据预处理的过程中对受试者的隐私信息进行了有效保护，并将在志愿者知情同意的前提下公开步态剪影数据以及对应的抑郁风险标注以推动这一领域的发展。
The statistical information of the dataset proposed in this paper is presented in Table~\ref{tab:DataStat_1}.
A comparison with existing datasets is illustrated in Table~\ref{tab:DataStat_2}.
In the existing dataset, the proposed dataset has the largest number of participants and gait sequences.
It also includes the highest number of camera views and attires conditions.
Due to privacy concerns, the majority of existing datasets cannot be openly accessed, which significantly impedes the progress in this field.
To address this issue, we have implemented effective privacy protection measures during the data preprocessing stage.
We will release gait silhouette data along with corresponding annotations of depression risk, but only with the informed consent of the volunteers.
This study aims to facilitate the advancement of research in this field while ensuring the protection of participants' privacy.

%%%%%%%%%%%%%%%%%%%%%%%%%%%%%%%%%%%%%%%%%%%%%%%%%%%%%%%%%%%%
%%%%%%%%%%%%%%         Depression Risk Recognition       %%%%%%%%%%%%%%%%%%
%%%%%%%%%%%%%%%%%%%%%%%%%%%%%%%%%%%%%%%%%%%%%%%%%%%%%%%%%%%%

\section{Depression Risk Recognition}
\label{sec:Method}
% ------------- Overview -------------- %

\subsection{Overview}

% 与一般的步态分析任务相比,基于步态的抑郁风险识别有两个显著的特点：
% 首先，与抑郁风险具有关联性的主要是动态特征，即人体行走过程中表现在时间维度上的步态特征，而人体的静态特征则与抑郁风险关联较小。
% 第二，与抑郁风险具有关联性的步态特征既可能表现为某个局部的细节特征，也可能涉及到整个人体。
% 基于以上两个特点，本文提出了一个以动态特征建模为核心的深度学习模型，并且在特征提取过程中将局部动态特征和全局动态特征进行了有效融合。

\begin{figure*}[t]
%\vspace{-0.1cm}
\begin{center}
\includegraphics[width=0.8\linewidth]{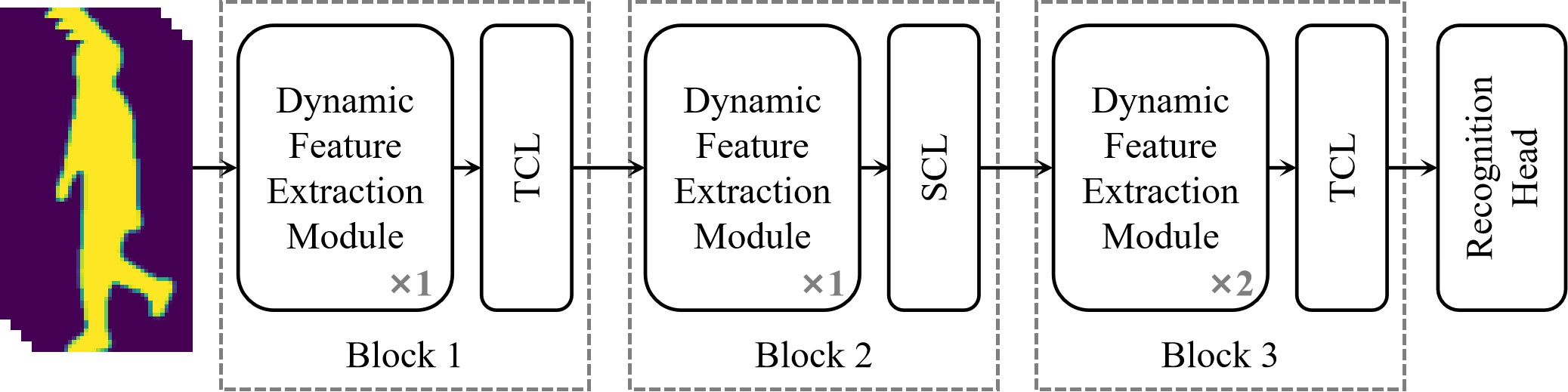}
\end{center}
   \caption{The framework of the proposed depression risk recognition model. The model consists of three kinds of modules: the dynamic feature extraction module, the feature compression module, and the recognition head. ``TCL" denotes the Temporal Compression Layer, ``SCL" denotes the Spatial Compression Layer.}
\label{fig:framework}
%\vspace{-0.6cm}
\end{figure*}

Compared to general gait analysis tasks, there are two notable characteristics in depression risk recognition based on gait analysis:
Firstly, the main features associated with depression risk are dynamic features, which refer to the temporal aspects of gait during the walking process\cite{sloman1982gait, lemke2000spatiotemporal, michalak2009embodiment, wang2020gait}.
Static features of the human body have less correlation with the depression risk, e.g., shape.
Secondly, gait features that are associated with depression risk can manifest as either local details or involve the entire body\cite{sloman1982gait, lemke2000spatiotemporal, michalak2009embodiment}.
Based on these two characteristics, this paper proposes a deep learning model with dynamic feature modeling as its core.
In the feature extraction process, both local dynamic features and global dynamic features are effectively integrated.

%%

% 这一深度学习模型由三类模块构成：动态特征提取模块、特征压缩模块以及recognition head，如图所示。
% 其中动态特征提取模块主要用于提取步态的动态特征，它由局部动态特征和全局动态特征两个分支构成。
% 特征压缩模块则主要对动态特征提取模块得到的动态特征进行压缩，进而得到更加紧致有效的表达。
% Recognition head用于将模型提取的动态特征映射至抑郁风险识别的任务空间。
This deep learning model consists of three kinds of modules: the dynamic feature extraction module, the feature compression module, and the recognition head, as shown in Fig~\ref{fig:framework}.
The dynamic feature extraction module is designed to extract the dynamic features of gait, and it consists of two branches: local dynamic feature branch and global dynamic feature branch.
The feature compression module compresses the dynamic features to obtain a more compact and effective representation.
The recognition head maps the final features to the task space of depression risk recognition.
%
% 接下来的小节将对这些模块逐一进行介绍
The following subsections will provide a detailed description to each of these modules.

%
% 画一个模型总体结构图

% ------------- Dynamic Feature Extraction -------------- %
\subsection{Dynamic Feature Extraction}

% 先讲一下动态特征的重要性，再讲一下为啥要局部全局分开，
% 然后讲局部：（1）很多人体分析都用分part的方法来建模局部，我们也是，将输入分为N个part，每个part分别提取动态特征，注意在讲卷积核的时候把卷积核在时间维度的用字母表示，用于控制动态特征的时间跨度
% 讲全局：提取动态特征，注意在讲卷积核的时候把卷积核在时间维度的用字母表示，用于控制动态特征的时间跨度
% 讲局部全局融合

% 画一个局部全局结构图

\begin{figure*}[t]
%\vspace{-0.1cm}
\begin{center}
\includegraphics[width=0.7\linewidth]{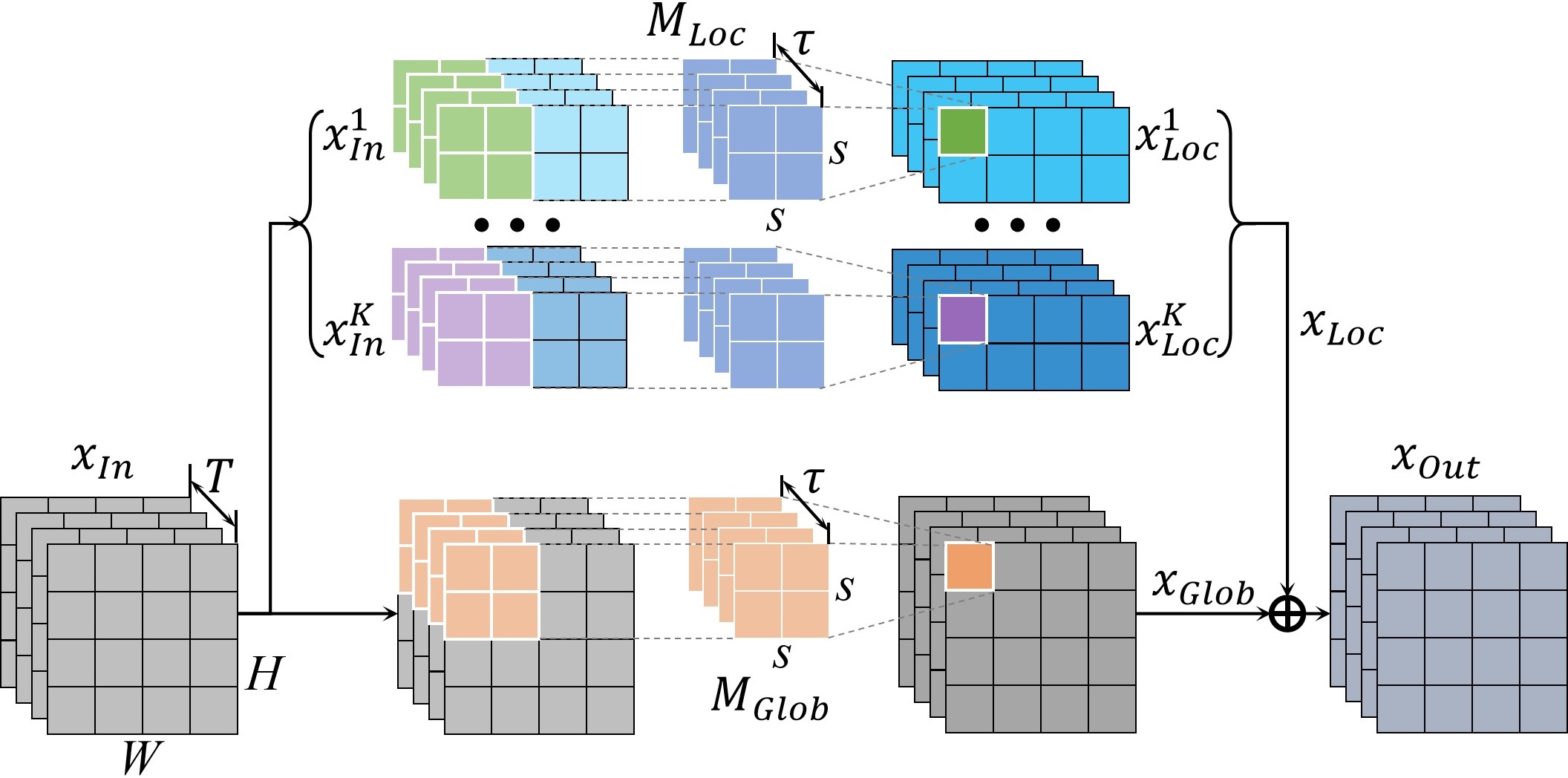}
\end{center}
   \caption{The proposed dynamic feature extraction module leverages 3D convolution to extract dynamic features. Additionally, it focuses on local dynamic features related to depression risk through the local feature branch, as well as global dynamic features through the global branch.}
\label{fig:two_branch}
%\vspace{-0.6cm}
\end{figure*}

% 现有的关于抑郁症与步态之间的关联性的研究表明，抑郁风险与步态之间的关联性主要体现在步态的动态特征上。
% 因此，有效提取和建模步态的动态特征是基于步态进行抑郁风险识别的关键。
% 为此，我们提出了一个新的动态特征提取模块。
% 此外，与抑郁风险关联的步态特征的表现形式是多样的，既可能涉及整个人体，也可能只表现在某个局部动作上。
% 因此，动态特征提取模块由两个分支组成，分别为局部特征分支与全局特征分支，分别用于提取局部和全局两类特征。
%%% 补充参考文献
Existing research on the association between depression and gait suggests that the correlation between depression risk and gait primarily manifests in the dynamic characteristics of gait.
Therefore, it is crucial to effectively extract and model the dynamic features of gait for the purpose of depression risk recognition based on gait.
To achieve this, we propose a novel dynamic feature extraction module.
Furthermore, the manifestation of gait features associated with depression risk varies, encompassing both the entire body and specific local movements.
Therefore, the dynamic feature extraction module consists of two branches: the local feature branch and the global feature branch, which are dedicated to extracting local and global features, respectively.

% 对于输入X，local feature branch首先将其在空间的垂直方向划分为K个part
When given input $x_{In} \in \mathbb{R}^{C \times T \times H \times W}$, where $C$ is the number of channels, $T$ is temporal scale of $x_{In}$, i.e., the length of the sequence, $H$ and $W$ are the spatial scale, the local feature branch initially divides it into $K$ parts in the vertical direction.
$x^k_{In} \in \mathbb{R}^{C \times T \times \frac{H}{K} \times W}$ is the $k$-th part, $k \in \{ 1, 2, ..., K \}$.
%
% 而后，一个局部3D卷积层被用于从每个part中提取动态特征：
Subsequently, a local 3D convolutional layer is employed to extract dynamic features from each part:
\begin{equation}
    x^k_{Loc} = M^{s \times s \times \tau}_{Loc}(x^k_{In})
\end{equation}
where $M^{s \times s \times \tau}_{Loc}$ is the local 3D convolutional layer, $s$ is the spatial kernel size of the 3D convolution operation, $\tau$ is the temporal kernel size.

%%

% 现有的步态分析研究工作大都在底层特征提取过程中采用2D卷积，这是由于动态特征对于一般的步态分析任务的重要性并不是很强。
% 但是由于动态特征对于抑郁风险识别而言十分关键，因此我们采用了3D卷积，使得模型能够直接在时间维度上获取步态模式。
% 参数t用于调节动态特征提取时的时间范围，当t较大时，模型更加倾向于捕捉较长时间的步态模式，当t较小时，模式则更倾向于捕捉短时间段内的步态模式。
Most existing studies on gait analysis primarily employ 2D convolutions for the process of low-level feature extraction.
However, considering the significance of dynamic features in depression risk recognition, we have opted for 3D convolutions, enabling the model to directly capture gait patterns in the temporal dimension.
The parameter $\tau$ is used to regulate the temporal range during dynamic feature extraction.
When $\tau$ is set to a higher value, the model tends to capture longer-term gait patterns, whereas a smaller $\tau$ value favors the capture of gait patterns within shorter time intervals.

%%

% 通过将输入划分为K个part，使得每个part之间的特征提取过程互不影响，让模型能够专注于当前的part即当前的局部区域进行动态特征提取。
% 因此，local feature branch能够关注与抑郁风险关联的局部特征。
By partitioning the input into K parts, we ensure that the feature extraction process between each part is independent, allowing the model to focus on a part, i.e., a local region of the input, for dynamic feature extraction.
This enables the local feature branch to concentrate on the localized features associated with depression risk.
%
% 在局部动态特征提取之后，K个局部特征被拼接在一起
After local feature extraction, the local feature maps of different parts are concatenated:
\begin{equation}
    x_{Loc} = Concat(x^1_{Loc}, x^2_{Loc}, ..., x^K_{Loc})
\end{equation}

%%

% global feature branch 在进行动态特征提取时也采用了3D卷积，但不会将输入划分为part：
The global feature branch also employs 3D convolutions for dynamic feature extraction, without the need to partition the input into parts.
\begin{equation}
    x_{Glob} = M^{s \times s \times \tau}_{Glob}(x_{In})
\end{equation}
where $M^{s \times s \times \tau}_{Glob}$ is the global 3D convolutional layer, $s$ is the spatial kernel size, $\tau$ is the temporal kernel size.
$\tau$ is also to regulate the temporal range during dynamic feature extraction.

%%

% dynamic feature extraction module 的输出由局部特征和全局特征融合得到
The output of the dynamic feature extraction module is obtained by merging local features and global features:
\begin{equation}
    x_{Out} = x_{Loc} + x_{Glob}
\end{equation}
%
% 通过融合局部特征和全局特征，最终完成dynamic feature extraction。
% 通过上述描述可以发现，dynamic feature extraction module通过应用3D卷积，实现了动态特征的有效提取，同时，通过局部和全局两个branch分别建模局部和全局特征。
% 这些有针对性的设计使得特征提取模型切合了抑郁风险识别任务的特点。
By integrating local and global features, the dynamic feature extraction process is accomplished ultimately.

From the aforementioned description, it can be found that the dynamic feature extraction module achieves efficient extraction of dynamic features through the application of 3D convolution.
Additionally, the local and global branches model the local and global features separately.
These targeted designs tailor the feature extraction model to the characteristics of depression risk recognition task.

% ------------- Feature Compression -------------- %
\subsection{Feature Compression}

% 画一个时间维度压缩图

% 由于步态数据的维度通常较高，而且存在大量的信息冗余，为了避免维度灾难、并提升特征表达的有效性，有必要在特征提取过程中进行特征压缩。
% 步态剪影数据的信息冗余主要来自两个方面；
% 首先是在时间上存在信息冗余，在步态数据采集过程中每秒采集30帧，相邻帧之间的时间间隔很短，有大量的信息是重复的。
% 其次，在空间上也存在信息冗余，剪影数据仅仅包含人体的轮廓信息，其信息密度远低于RGB数据，但其维度却与RGB数据一致，因此在空间上也存在大量的信息冗余。
% 基于上述洞见，我们在本文中提出了两种特征压缩层，一种在时间上进行特征压缩，一种在空间上对特征进行压缩。
Due to the typically high dimensionality of gait data and the presence of significant information redundancy, it is necessary to perform feature compression during the feature extraction process to mitigate the curse of dimensionality and enhance the effectiveness of feature representation.
Information redundancy in gait silhouette data primarily stems from two aspects.
Firstly, there is temporal redundancy as gait data is collected at a rate of 30 frames per second, resulting in short time intervals between adjacent frames and a substantial amount of repeated information.
Secondly, there is spatial redundancy.
Silhouette data only captures the contour information of the human body, which has lower information density compared to RGB data.
However, the dimensionality of silhouette data remains the same as RGB data, leading to a significant amount of spatial information redundancy.
Building upon the above insights, we propose two types of feature compression layers in this paper: one for temporal compression and the other for spatial compression of features.

%%

% 对于feature map x， 为了方便表示，将时间维度意外的维度略去，一种简单直接的特征压缩方法是线性压缩
%\noindent\emph{\textbf{Temporal compression layer:}}

%%

\begin{figure}[t]
%\vspace{-0.1cm}
\begin{center}
\includegraphics[width=0.7\linewidth]{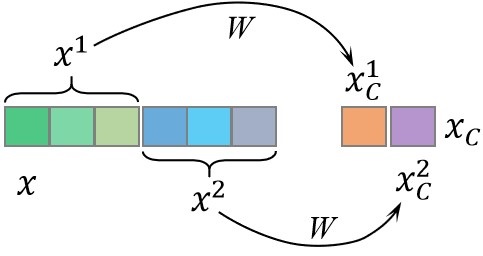}
\end{center}
   \caption{Temporal Compression Layer (TCL) is proposed to accomplish feature compression in the temporal dimension. The input is grouped along the temporal dimension and perform separate compression for each group.}
\label{fig:TCL}
%\vspace{-0.6cm}
\end{figure}

Temporal Compression Layer (TCL) is proposed to accomplish feature compression in the temporal dimension, as shown in Fig.~\ref{fig:TCL}.
For the feature map $x \in \mathbb{R}^{C \times T \times H \times W}$, dimensions other than the temporal dimension are omitted for the sake of simplicity: $x \in \mathbb{R}^T$.
A straightforward method for feature compression is linear compression:
\begin{equation}
    x_C = W x
\end{equation}
where $x_C \in \mathbb{R}^N$ is the compressed feature, $N$ is the temporal dimension after compression, $W\in \mathbb{R}^{N \times T}$ is the compression parameter.
%
% 但是这种方式会导致模型的参数量大为增加，不利于模型的训练和效率。
% 为此，我们提出在时间维度上对输入进行分组，每组分别压缩
%
However, this approach would result in a significant increase in the number of model parameters, which is not conducive to model training and efficiency.
To address this issue, we propose to partition the input along the temporal dimension and perform separate compression for each group:
\begin{equation}
    x_C^i = W^i x^i
\end{equation}
where $x^i \in \mathbb{R}^{\frac{T}{N}}$ is the $i$-th group, $i \in \{1, 2, ..., N\}$, $W^i \in \mathbb{R}^{1 \times \frac{T}{N}}$ is the compression parameter, $x_C^i \in \mathbb{R}^{1\times 1}$ is the $i$-th compressed feature.
%
% 进一步地，采取权重共享策略：
Furthermore, we adopt a weight sharing strategy:
:
\begin{equation}
    \forall i \in \{ 1, 2, ..., N \}, W^i = W
\end{equation}
%
% 最后将压缩后的特征在时间维度上拼接得到压缩后的特征：
Finally, the compressed features are concatenated along the temporal dimension to obtain the compressed feature:
\begin{equation}
    x_C = Concat(x_C^1, x_C^2, ..., x_C^N)
\end{equation}
where $x_C \in \mathbb{R}^N$.
% 上述时间特征压缩方法的合理性来自于与步态数据以及抑郁风险相关特征的洞见：即步态在时间上是连续的，相邻帧之间的差异很小，因此我们将在时间上相邻的输入划分为同一组进行压缩。
% 与直接进行线性压缩相比，所提出的方法的参数量大大下降，由TN下降至N。
The rationale behind the aforementioned method of compressing temporal features is derived from insights into gait data and related features associated with depression risk.
Specifically, it is observed that gait is continuous over time and there is minimal variation between adjacent frames.
Hence, we partition the inputs that are temporally adjacent into the same group for compression.
Compared to directly performing linear compression, the proposed method significantly reduces the number of parameters, decreasing it from $T\cdot N$ to $T/N$.

% 空间压缩

Spatial Compression Layer (SCL) is employed to accomplish feature compression in the spatial dimensions.
%
% 空间特征压缩是深度学习模型中常见的操作，max pooling是应用最多的空间压缩方法。
% 因此，在所提出的模型中，max pooling被用于实施空间压缩。
Spatial feature compression is a common operation in deep learning models, and max pooling is the most commonly used spatial compression method.
Therefore, in the proposed model, max pooling is utilized to implement spatial compression.
%%

% ------------- Recognition Head -------------- %
\subsection{Recognition Head}

% 在完成特征提取之后，提取得到的特征f被送入recognition head以对输入样本的抑郁风险进行识别。
% 本文首先通过recognition head对是否存在抑郁风险进行分类：
Upon completion of feature extraction, the extracted feature $f$ is fed into the recognition head for the depression risk recognition.
The recognition head initially performs classification to determine the presence of depression risk:
\begin{equation}
    y = M_{Head}(f)
\end{equation}
where $M_{Head}$ is a fully connected layer, $y\in \mathbb{R}^{2}$ is the output of $M_{Head}$.
% 
% 在训练过程中通过交叉熵损失函数进行训练：
During the training process, the model is trained using the cross-entropy loss function:
\begin{equation}
    L_{CE} = -\sum_{i=1}^N y^*_i ln \frac{e^{y_i}}{\sum_j e^{y_j}}
\end{equation}
where $N$ is the number of classes and $N=2$, $y^*$ is the depression risk label, when a sample is at risk of depression $y^*_1=1,~y^*_2=0$, and when a sample is not at risk of depression $y^*_1=0,~y^*_2=1$.

%%

% 与此同时，抑郁风险识别又不能简单地概括为一个二分类问题，这是由于受试者的心理状态具有独特性，即使对于两个都具有抑郁风险的受试者而言，他们的心理状态也可能存在很大差别。
% 因此，我们在上述分类损失之外，我们采用triplet loss对模型提取得到的特征进行约束：
Meanwhile, the depression risk recognition cannot be simply generalized as a binary classification problem.
This is due to the uniqueness of individuals' psychological states, where even for two participants both at risk of depression, their psychological states may vary significantly.
Hence, in addition to the aforementioned classification loss, we employ triplet loss to constrain the features extracted by the model:
\begin{equation}
    L_{Tri} = max(D(f_A, f_P) - D(f_A, f_N) + m, ~ 0)
    \label{eq:triplet}
\end{equation}
%
% 其中D是特征空间中的距离度量函数，a是anchor，p是正样本，即p与a对应于相同的受试者，n是负样本，即n与a对应于不同的受试者，
where $D(\cdot, \cdot)$ represents the distance metric function in the feature space, $f_A$ denotes the anchor, $f_P$ denotes the feature of the same subject as $f_A$, and $f_N$ denotes the feature of a different subject than $f_A$, $m$ is the distance margin.
%
% 这一损失函数要求对于同一个人而言，模型提取到的抑郁风险特征要保持相对稳定和一致，同时不同人的特征要有所区别。
This loss function requires that for the same individual, the extracted depression risk features by the model should remain relatively stable and consistent, while features of different individuals should exhibit distinctiveness.

%%

% 模型的训练目标即由上述两个损失函数构成：
The training objective of the model is composed of the aforementioned two loss functions:
\begin{equation}
    L = L_{CE} + L_{Tri}
\end{equation}

% ------------- Implementation details -------------- %

\subsection{Implementation details}

% 本小节介绍所提出的抑郁风险识别模型的实现细节以及训练和测试设置。
This subsection introduces the implementation details of the proposed depression risk recognition model, as well as the training and testing configurations.

%%

% 剪影数据的空间维度被归一化至64*44，在训练过程中，从输入序列中随机抽取连续的60帧作为输入，当序列不足60帧时，通过重复序列补足60帧，测试时则将测试数据的所有帧均送入模型进行测试。
% 训练时采用Adam优化器，的初始学习率为10-4，weight decay=5e-4,batch size为64，总共训练8万步，在第7万步时学习率衰减为原来的十分之一。
The spatial dimensions of the silhouette data are normalized to $64\times 44$.
During the training process, a random sequence of 60 consecutive frames is extracted as the input sequence.
If the sequence is less than 60 frames, it is padded by repeating the sequence until it reaches 60 frames.
During testing, all frames of the test data are inputted into the model for testing.
For training, the Adam optimizer is used with an initial learning rate of 10-4 and a weight decay of 5e-4.
The batch size is set to 64, and a total of 80,000 steps are performed. The learning rate is decayed to one-tenth of the original value at the 70,000th step.

% 三个block的通道数设置分别为32，64，128.
% 动态特征提取模块中的局部特征分支将输入划分为16个part，局部和全局3D卷积层的空间kernel size均为3，时间kernel size均为3。
% block1 中的时间特征压缩层在时间维度将特征分为20组进行压缩，block 3中的时间特征压缩层则只划分一个组进行压缩。
% 公式中的距离度量函数D采用欧式距离，m为0.2
The channel of the three blocks are set as 32, 64, and 128 respectively.
In the dynamic feature extraction module, the local feature branch divides the input into 16 parts.
Both the spatial kernel size of the local and global 3D convolutional layers are set to 3, and the temporal kernel size are set to 5.
In block 1, the temporal feature compression layer divides the features into 20 groups for compression in the temporal dimension.
In block 3, the time feature compression layer divides the features into one group for compression.
Euclidean distance is utilized as the distance metric function $D(\cdot, \cdot)$ in Eq.\ref{eq:triplet}, with $m$ set to 0.2.

%%%%%%%%%%%%%%%%%%%%%%%%%%%%%%%%%%%%%%%%%%%%%%%%%%%%%%%%%%%%
%%%%%%%%%%%%%%         Experiments       %%%%%%%%%%%%%%%%%%
%%%%%%%%%%%%%%%%%%%%%%%%%%%%%%%%%%%%%%%%%%%%%%%%%%%%%%%%%%%%

\section{Experiments}

\label{sec:Experiments}
%%
% 为了验证所提出的抑郁风险识别的有效性，我们在采集的步态数据集上进行了定量实验。
% 本节首先对实验设置进行介绍，而后对所提出的模型进行抑郁风险识别的有效性就行了定量验证。
% 之后，我们分别就步态视角对模型预测的影响以及抑郁风险分级与模型预测的关系给出了定量分析。
% 最后我们对动态特征建模以及损失函数进行了消融实验。
In order to validate the effectiveness of the proposed depression risk recognition model, we conducted quantitative experiments on the collected gait dataset.
This section first introduces the experimental protocols, and then quantitatively validates the effectiveness of the proposed model for depression risk recognition.
Subsequently, we provide a quantitative analysis on the impact of views on the model's predictions, as well as the relationship between grading of depression risk and the model's predictions.
Finally, we conducted ablation experiments on the dynamic feature modeling and loss functions.

% ------------- Protocols -------------- %

\subsection{Protocols}

%%

% 在实验过程中，采集的步态数据集被分为训练集和测试集，训练集合包含831个受试者，测试集包含396个受试者。
% 抑郁风险识别模型在训练集上进行训练，并在测试集上对其识别性能进行定量验证。
% 每个受试者的步态数据都被划分为走向摄像机以及远离摄像机的两个序列分别进行训练和测试。
% 在测试过程中，识别模型输出的具有抑郁风险的类别的概率被视为具有抑郁风险的概率。
During the experiment, the collected gait dataset is divided into a training set and a test set.
The training set consisted of 831 subjects, while the test set consisted of 396 subjects.
The training set and test set do not have any overlapping subjects.
The proposed depression risk recognition model is trained on the training set and its recognition performance is quantitatively validated on the test set.
Each subject's gait data is divided into two sequences: walking towards the cameras and walking away from the cameras during training and testing.
During the testing process, the probability of the class with depression risk, as outputted by the recognition model, is considered as the probability of having depression risk.
%

% 在对模型的识别性能进行评价时，我们采用以下几个性能指标：
The model's performance is evaluated using the following performance metrics:

\begin{equation}
    Acc = \frac{TP+TN}{TP+FP+TN+FN}\times 100\%
\end{equation}
where $TP$ is the number of true positive samples, $TN$ is the number of true negative samples, $FP$ is the number of false positive samples, $FN$ is the false negative samples.
%
% Acc是一种直观的评价指标，直接计算识别正确的样本所占的比例
Accuracy ($Acc$) is an intuitive evaluation metric that directly calculates the proportion of correctly recognized samples.

%%

% Precision
\begin{equation}
    Prec = \frac{TP}{TP+FP} \times 100\%
\end{equation}
%
% precision 
When the cost of incorrectly predicting negative samples as positive ones ($FP$) is high, precision ($Prec$) becomes a crucial metric.

% Recall rate:
\begin{equation}
    Recall = \frac{TP}{TP+FN}\times 100\%
\end{equation}
Conversely, when the cost of predicting positive samples as negative ones ($FN$) is high, recall rate ($Recall$) becomes an important metric.

% F1 score:
\begin{equation}
    F1 = \frac{2 \times Prec \times Recall}{Prec+Recall}
\end{equation}
%
% F1是一个综合了Precision和recall rate的评价指标，能够比较综合地评价模型的识别性能
$F1$ score is an evaluation metric that combines both precision and recall rate, providing a comprehensive assessment of a model's recognition performance.

%

% AUC
% 此外，Receiver operating characteristics (ROC) curve 的曲线下面积（AUC）也被用于定量评价模型的识别性能，这一指标也是一项综合性指标，并且具有不受阈值选择影响的优点。
In addition, the area under the curve (AUC) of the Receiver Operating Characteristics (ROC) curve is also used to quantitatively evaluate the recognition performance of the model.
This metric is comprehensive and has the advantage of being unaffected by threshold selection.

% 由于测试集中的正样本和负样本的数量基本均衡，因此，在进行性能评价时，首先根据模型的预测结果选择一个使得Accuracy最优的抑郁风险识别阈值，而后在此阈值的基础上计算上述性能指标。
% 

% ------------- Depression Risk Recognition -------------- %
\subsection{Depression Risk Recognition}

% 所提出的抑郁风险识别模型在所采集的数据集上的性能如表所示。
% 可以看到，所提出的模型的几个指标都分布在80%左右，这表明模型已经能够将大部分的正负样本进行区分，从而得到一个差强人意的结果。
% 这些实验结果进一步验证了基于步态对抑郁风险进行识别的可行性，同时也验证了所提出的深度学习模型的有效性。
The performance of the proposed depression risk recognition model on the collected dataset is shown in Table~\ref{tab:conv5}.
As can be observed, the metrics of the proposed model are distributed around 80\%, indicating that the model is able to distinguish between most positive and negative samples, resulting in a passable outcome.
These experimental results further validate the feasibility of depression risk recognition based on gait and also confirm the effectiveness of the proposed deep learning model.

\begin{table}[t]
\begin{center}
\caption{The performance of the proposed depression risk recognition model on the collected dataset.}
\label{tab:conv5}
\setlength{\tabcolsep}{4mm}
{
\begin{tabular}{c|c|c|c|c}
\hline
\hline
        \bf{Acc} $\uparrow$ & \bf{Prec}$\uparrow$ & \bf{Recall}$\uparrow$ & \bf{F1}$\uparrow$ & \bf{AUC}$\uparrow$ \\
\hline
         79.41\% & 76.71\%   & 82.94\%     & 79.70\% & 0.8019   \\

\hline
\hline
\end{tabular}}
\end{center}
\end{table}

\begin{figure}[t]
%\vspace{-0.1cm}
\begin{center}
\includegraphics[width=0.9\linewidth]{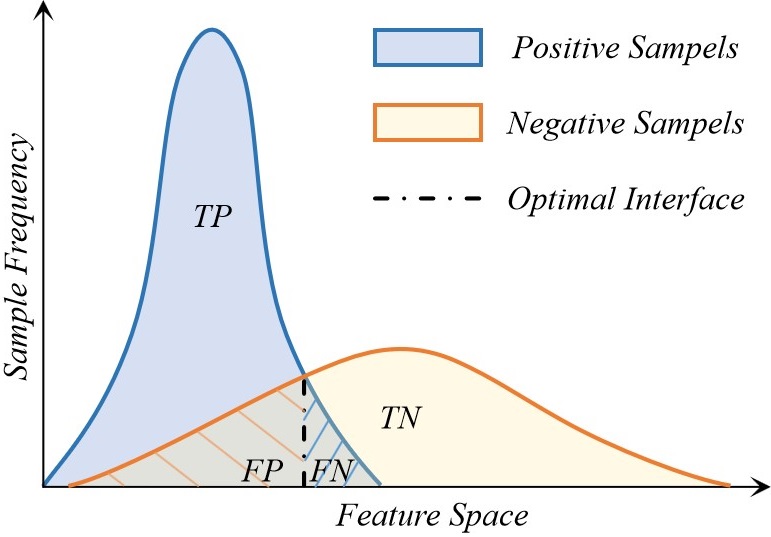}
\end{center}
   \caption{Diagram illustrating the distribution of samples in the feature space. For the task of depression risk recognition, positive samples exhibit a tendency to cluster together in the feature space, while negative samples are more dispersed. For the purpose of convenience, the feature space is simplified to a one-dimensional space. This clustering pattern leads to a higher number of false positive samples ($FP$) compared to false negative samples ($FN$), resulting in a recall rate that is greater than the precision rate.}
\label{fig:PosNeg}
%\vspace{-0.6cm}
\end{figure}

%%

% 从实验结果中可以发现，模型在进行抑郁风险识别时，正样本的召回率明显高于精度。
% 这一现象说明模型提取到的特征更加善于将具有抑郁风险的正样本聚集在一起，相对而言不具有抑郁风险的负样本的则更加分散。
% 这是符合直觉的，即具有抑郁特征的正样本具有共有的特征表现，使得正样本之间具有较强的共性。
% 而不具有抑郁风险的负样本则不一定存在相同的特征，他们彼此之间的共性要明显少于正样本，从而导致其在特征空间中的分布更加分散。
% 如图所示意的，在特征空间中，正样本更加聚集而负样本更加分散。
% 此时模型能够学习到的最优分类界面会导致false positive samples的数量大于false negative samples 的数量，从而导致在Recall rate大于precision。
From the experimental results, it can be observed that the recall of positive samples in the depression risk recognition is significantly higher than the precision.
This phenomenon indicates that the features extracted by the model are more adept at clustering positive samples with depression risk together, while negative samples without depression risk are relatively scattered.
This is intuitive as positive samples with depression-related characteristics exhibit common feature representations, resulting in strong similarities between them.
On the other hand, negative samples without depression risk may not have the same set of features. 
The similarities among negative samples are much less pronounced compared to positive samples, resulting in a more dispersed distribution in the feature space.

As depicted in Fig.~\ref{fig:PosNeg}, positive samples tend to cluster together in the feature space, while negative samples are more dispersed.
Consequently, the optimal classification interface that the model can learn in this scenario would result in a larger number of false positive samples ($FP$) compared to false negative samples ($FN$), leading to a higher recall rate than precision rate.

% ------------- Influence of Views -------------- %
\subsection{Influence of Views}

%%

% 在一般的步态分析任务中，拍摄视角是一个重要的影响因素。
% 这是因为在不同的视角下所观察到的步态区别较大，步态的某些特征可能在某些视角下很容易观察，而在另一些视角下则难以获取。
% 因此，有必要针对视角对抑郁风险识别的影响进行进一步探究。
In general gait analysis tasks, the view of camera plays a crucial role as an influencing factor.
This is because there can be significant differences in the observed gait patterns across different views.
Certain gait features may be easily observable from specific views, while they may be challenging to capture from other views.
Therefore, it is necessary to further explore the impact of views on depression risk recognition.

\begin{table}[t]
\begin{center}
\caption{The performance of the proposed depression risk recognition model on different views. The recognition models based on each separate view of camera is inferior to the performance achieved with all views combined. However, it is worth noting that the difference in performance between the individual views of camera and the overall views is not significant.}
\label{tab:views}
\setlength{\tabcolsep}{3.1mm}
{
\begin{tabular}{c|c|c|c|c|c}
\hline
\hline
    \bf{View}  &  \bf{Acc} $\uparrow$ & \bf{Prec}$\uparrow$ & \bf{Recall}$\uparrow$ & \bf{F1}$\uparrow$ & \bf{AUC}$\uparrow$ \\
\hline
      \bf{\# 1}    &  77.65\% & 74.57\%   & 82.07\%     & 78.14\% & 0.7989   \\
      \bf{\# 2}    &  78.18\% & 76.69\%   & 82.70\%     & 79.58\% & 0.8008   \\
%\hline
      \bf{\# 3}    &  77.70\% & 75.84\%   & 80.60\%     & 78.15\% & 0.7810   \\
      \bf{\# 4}    &  78.86\% & 76.31\%   & 82.69\%     & 79.37\% & 0.7920   \\
%\hline
      \bf{\# 5}    &  77.82\% & 75.19\%   & 80.21\%     & 77.62\% & 0.7937   \\
      \bf{\# 6}    &  79.07\% & 76.38\%   & 81.72\%     & 78.96\% & 0.8013   \\
\hline
%\hline
      \bf{All}     &  \bf{79.41}\% & \bf{76.71}\%   & \bf{82.94}\%     & \bf{79.70}\% & \bf{0.8019}   \\

\hline
\hline
\end{tabular}}
\end{center}
\end{table}

% 为此，我们将训练集和测试集均按照视角划分为6个子集，每个子集中只包含一种视角下的步态数据。
% 将所提出的抑郁风险识别模型分别在这6个子集上进行训练和测试。
% 实验结果如表所示，其中视角的编号与图1保持一致
To accomplish this, we divide the training and testing datasets into six subsets based on views of camera.
Each subset exclusively contains gait data captured from a specific view.
We then proceed to train and test the proposed depression risk recognition model on these six subsets independently.
The experimental results are presented in Table~\ref{tab:views}, with the angle numbering consistent with Fig.~\ref{fig:gait_collect}.

% 从中我们可以发现，所有单独的视角下的识别性能均不如全视角下的性能。
% 这说明每一个单独视角所捕捉到的抑郁风险相关的特征都不如全视角全面，不同视角之间存在一定的信息互补，即不同视角观察到了抑郁风险相关步态特征的不同方面。

% 但是于此同时应当注意到，单个视角下的识别性能与全视角下的性能差距并不明显。
% 这说明对于抑郁风险识别任务而言，视角之间互补的有效信息是有限的，大部分的抑郁相关特征在所有视角下均能观察到。
% 而空间维度的静态特征受视角的影响较大，如摆臂角度，时间维度的动态特征则受视角的影响较小，在不同视角均能捕捉到，例如摆臂频率。
% 这再次表明，抑郁相关的步态特征更多的是表现在时间维度的动态特征，而非表现在空间维度的静态特征。
From these results, we can observe that the performance of the recognition models based on each separate view of camera is inferior to the performance achieved with all views combined.
This indicates that the features related to depression risk captured by each individual view are not as comprehensive as those captured by the complete set of views.
Different angles provide complementary information of gait features related to depression risk.
However, it is important to note that the difference in performance between the individual views of camera and the overall views is not significant.
This suggests that the effective complementary information between different views is limited for the task of depression risk identification.
Most of the depression-related features can be observed in all views.
Static features in the spatial dimension, such as arm swing angles, are more influenced by view of camera, while dynamic features in the temporal dimension can be captured in multi-views, such as arm swing frequency.
This further verifies that the gait features related to depression risk are primarily manifested in dynamic features in the temporal dimension rather than static features in the spatial dimension.

%%

% 此外，还可以观察到，距离地面较高的视角（#1，#3，#5）与其对应的较低的视角（#2，#4，#6）相比，性能较差。
% 表明较低的视角更有利于抑郁相关特征的捕捉，这是由于较高的视角会导致人体的自遮挡现象更加严重，尤其不利于腿部动作的观察。
Additionally, we can also observe that the performance of the models captured from higher views (\# 1, \# 3, \# 5) is inferior compared to the corresponding lower views (\# 2, \# 4, \# 6).
This indicates that lower views of camera are more favorable for capturing depression-related features.
This is because higher views of camera contribute to a more severe self-occlusion phenomenon in the human body, particularly impeding the observation of leg movements.

% ------------- Relationship between Grading of Depression Risk and Model Predictions -------------- %
\subsection{Relationship between Grading of Depression Risk and Model Predictions}

%%

% 抑郁症的分级是抑郁症研究的一个重要方面。
% 本小节探究所提出的模型对抑郁风险的预测与样本的抑郁风险分级之间的关系。
% 为此，我们依据SDS和PHQ-9两个量表的分数，对受试者的抑郁风险进行了分级。
% 对于SDS来说，分数的大于等于70为有严重的抑郁风险，分数小于70大于58为有中度抑郁风险。
% 对于PHQ-9来说，分数的大于等于15为有严重的抑郁风险，分数小于15大于8为有中度抑郁风险。
The grading of depression is an integral aspect of depression research.
This subsection explores the relationship between the predictions made by the proposed model and the grading of depression risk.
To achieve this, the participants' depression risk is classified based on the scores obtained from the SDS and PHQ-9 scales.
For the SDS, a score of 70 or above indicates a severe risk of depression, while a score below 70 but above 58 indicates a moderate risk of depression.
As for the PHQ-9, a score of 15 or above indicates a severe risk of depression, while a score below 15 but above 8 indicates a moderate risk of depression.

\begin{figure}[t]
%\vspace{-0.1cm}
\begin{center}
\includegraphics[width=0.9\linewidth]{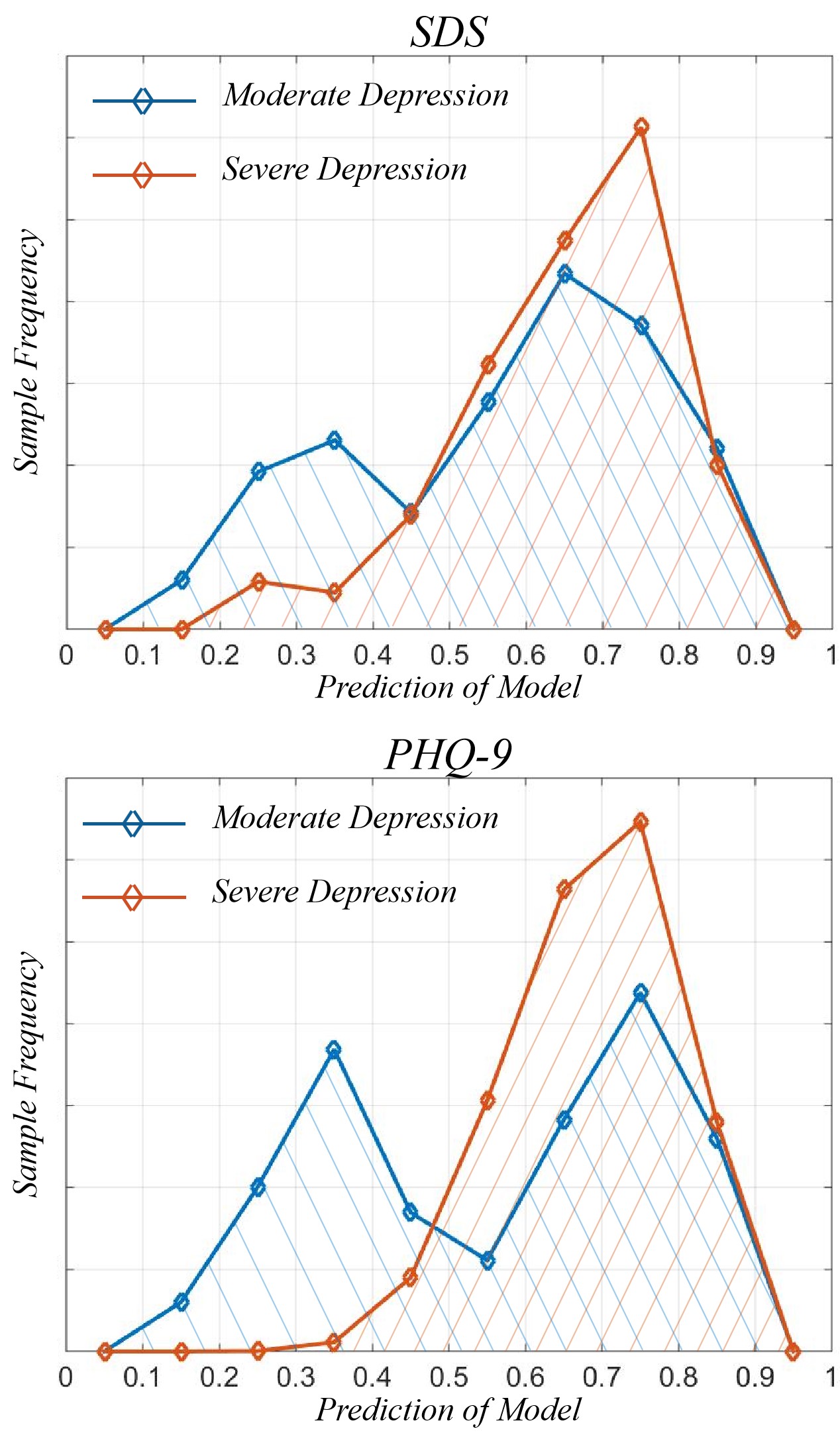}
\end{center}
   \caption{The depression probabilities predicted by the model of the samples with moderate and severe depression risk.}
\label{fig:SDS}
%\vspace{-0.6cm}
\end{figure}

%%

% 中度抑郁风险和重度抑郁风险的样本的模型输出被分别统计。
% 模型输出的抑郁概率被划分为10个区间进行统计，SDS和PHQ-9的模型输出分布如图所示。
% 从图中可以发现，具有重度抑郁风险的样本对应的模型预测较为集中的分布在0.7左右。
% 相反，具有中度抑郁风险的样本则呈现双峰分布，其中一部分与具有重度抑郁风险的样本具有相似的分布，而另一部分对应的模型预测则集中在更低的概率附近。
% 由于所提出的深度学习模型在训练过程中并没有依据抑郁症的分级进行针对性的训练，因此这一现象暗示模型在学习过程中捕捉到的重度抑郁风险相关的步态特征和中度抑郁风险相关的步态特征只有一部分是共享的，而没有出现这部分特征的中度风险样本就与具有这部分特征的样本在预测上出现了差异，从而导致模型对与具有中度抑郁风险的样本的预测呈现双峰。
The predictions made by the proposed model for samples with moderate and severe depression risks were separately analyzed.
The depression probabilities predicted by the model are categorized into 10 intervals, as shown in Fig.~\ref{fig:SDS}.
From the figures, it can be observed that samples with a severe depression risk exhibit a more concentrated distribution around 0.7 in the model predictions.
In contrast, samples with a moderate depression risk show a bimodal distribution.
One part of the distribution closely resembles that of samples with severe depression risk, while the other part is centered around lower probabilities.
Since the proposed deep learning model was not explicitly trained to consider the grading of depression during its training process, this phenomenon suggests that only a portion of the features related to severe and moderate depression risks are shared in the proposed deep learning model.
As a result, samples with moderate risk that lack these shared features exhibit differences in the predictions compared to samples with these features, leading to the observed bimodal distribution in the model's predictions for samples with moderate depression risk.

% ------------- Ablation Study -------------- %
\subsection{Ablation Study}

% Dynamic Feature Modeling %
\subsubsection{Dynamic Feature Modeling}

%%

% 本文所提出的抑郁风险识别模型的关键是建模时间维度的动态特征。
% 为此，能够在时间维度进行特征表达的3D卷积模型被作为搭建该深度学习模型的基础操作。
% 在模型搭建过程中，3D卷积核的temporal kernel size t 是重要的超参数。
% 为了探究3D卷积核的temporal kernel size对抑郁风险识别的影响，我们采用三种不同大小的temporal kernel size进行了实验，分别是3，5，7。
The crux of the proposed model for depression risk recognition lies in effectively modeling the dynamic features in the temporal dimension.
In order to achieve this, we employed the 3D convolutional operation that enables us to capture temporal patterns as the fundamental building block of our deep learning architecture.
Throughout the construction process of our model, the selection of the temporal kernel size (t) for the 3D convolution operation emerged as a crucial hyperparameter.
It is necessary to investigate the influence of different temporal kernel sizes on the performance of the proposed depression risk recognition model.
In light of this, we conducted experiments employing three distinct temporal kernel sizes: 3, 5, and 7.

\begin{table}[t]
\begin{center}
\caption{The influence of the temporal kernel size of the 3D convolutional kernel on the recognition performance of the proposed model. The bold font signifies the optimal performance, while the underlined font denotes the second-best performance. It is evident that as the temporal kernel size increases, the overall performance of the model improves. And the performance gains attributed primarily to the increase in recall rates.}
\label{tab:tau}
\setlength{\tabcolsep}{3.1mm}
{
\begin{tabular}{c|c|c|c|c|c}
\hline
\hline
      &  \bf{Acc} $\uparrow$ & \bf{Prec}$\uparrow$ & \bf{Recall}$\uparrow$ & \bf{F1}$\uparrow$ & \bf{AUC}$\uparrow$ \\
\hline
      $\tau$ =3    &  79.17\%    & \bf{76.72}\%   & 82.22\%     & 78.37\% & 0.8018   \\
      
      $\tau$ =5    &  \underline{79.41}\%     & \underline{76.71}\%   & \underline{82.94}\%     & \underline{79.70}\% & \underline{0.8019}   \\

      $\tau$ =7    &  \bf{79.51}\% & 76.69\%   & \bf{83.33}\%     & \bf{78.84}\% & \bf{0.8037}   \\
      
\hline
\hline
\end{tabular}}
\end{center}
\end{table}

% 实验结果如表所示。
% 从实验结果中可以发现，随着temporal kernel size的增加，模型的总体性能总体性能随之增加，这一点可以从Acc，F1 score以及AUC这三个综合性的指标上观察得到。
% 进一步观察precision和recall rate可以发现，三者的precision差异很小，temporal kernel size带来的性能增益主要来自recall rate的增加。
% 这说明增加temporal kernel size主要提升了具有抑郁风险的正样本的表达效果，使得false negative （FN） samples减少了。
% 这是由于在模型深度不变的条件下，增加temporal kernel size能够增加模型中神经元在时间维度的感受野，使得模型能够更加有效地建模抑郁相关的动态特征的，从而更好地聚合具有抑郁风险的正样本，提升recall rate。
The experimental results are shown in Table~\ref{tab:tau}.
Upon analysis of the experimental outcomes, a conspicuous pattern emerges: an escalating temporal kernel size propels an ascendant trajectory in the overall efficacy of the model.
This salient observation is corroborated by the holistic evaluation metrics of Accuracy, F1 score, and AUC.
Furthermore, probing into the precision and recall rates divulges an inconsequential variance in precision, indicating that the performance enhancement engendered by the temporal kernel size alteration principally stems from the augmentation of recall rates.
This indicates that increasing the temporal kernel size predominantly enhances the representation of positive samples associated with depression risk, effectively reducing the number of false negative (FN) samples.
This can be attributed to the fact that, under the constraint of unchanged model depth, increasing the temporal kernel size expands the receptive field of neurons in the temporal dimension.
Consequently, the model can more effectively capture the depression-related dynamic features, resulting in better discrimination of positive samples with depression risk and an overall improvement in recall rate.

% Triplet Loss %
\subsubsection{Triplet Loss}

%%

% 在模型训练的过程中，triplet loss被用于约束模型提取得到的动态特征，以使得样本在特征空间中的分布更加有利于抑郁风险识别。
% 为了验证triplet loss的作用，我们在没有triplet loss作为约束的条件下进行训练作为对比。
During the process of model training, triplet loss, $L_{Tri}$ in Eq.~\ref{eq:triplet}, is employed to constrain the dynamic features extracted by the model, aiming to shape the distribution of samples in the feature space to be more conducive to depression risk identification.
In order to validate the effect of triplet loss, we conducted training without the constraint of triplet loss as a comparison.

\begin{table}[t]
\begin{center}
\caption{Ablation study on the triplet loss. The experiments indicate that the absence of triplet loss leads to significant performance degradation in the model.}
\label{tab:triplet}
\setlength{\tabcolsep}{2.6mm}
{
\begin{tabular}{c|c|c|c|c|c}
\hline
\hline
      &  \bf{Acc} $\uparrow$ & \bf{Prec}$\uparrow$ & \bf{Recall}$\uparrow$ & \bf{F1}$\uparrow$ & \bf{AUC}$\uparrow$ \\
\hline
      w/o $L_{Tri}$    &  50.02\%    & 48.72\%   & 51.18\%   & 49.36\% & 0.5001   \\
\hline     
      with $L_{Tri}$   &  \bf{79.41}\%    & \bf{76.71}\%   & \bf{82.94}\%     & \bf{79.70}\% & \bf{0.8019}   \\

\hline
\hline
\end{tabular}}
\end{center}
\end{table}

% 实验结果如表所示。
% 实验表明，没有triplet loss对模型造成的性能损失是严重的。
% 单纯依靠分类损失函数进行训练，模型在训练集上难以收敛，在测试集上几乎失去了抑郁风险识别能力。
% 这一现象表明，抑郁风险识别并不是一个简单的二分类问题，简单直接地使用分类损失函数行不通。
The experimental results are shown in Table~\ref{tab:triplet}.
The experiments indicate that the absence of triplet loss leads to significant performance degradation in the model.
Relying solely on the classification loss function for training, the model struggles to converge on the training set and almost loses its ability to recognize depression risk on the test set.
This phenomenon suggests that depression risk recognition is not a simple binary classification problem, and the straightforward use of a classification loss function is not viable.

%%

% triplet loss的作用是将来自同一个受试者的不同序列在特征空间中的距离拉近，这是合理的因为同一个受试者的不同序列的抑郁风险应当是一致的。
% 而当没有triplet loss时模型便难以收敛，这说明同一受试者的不同序列的抑郁风险相关的表现差异很大，导致模型在没有直接引导的情况下难以训练。
% 这种差异主要来自于步态这一特殊的数据形式：不同视角、不同着装的步态数据的特征表现差异很大，这是我们基于步态进行抑郁风险识别时需要克服的重要障碍。
The purpose of incorporating triplet loss is to minimize the inter-sequence distance within the feature space for samples originating from the same subject.
This approach is justified by the expectation that different sequences from the same subject should exhibit consistent levels of depression risk.
In contrast, the absence of triplet loss impedes model convergence, indicating substantial disparities in the manifestation of depression risk across distinct sequences from the same subject.
Consequently, the model struggles to train effectively in the absence of explicit guidance.
These discrepancies primarily arise from the distinctive nature of gait data, characterized by notable variations in feature expression due to dissimilar views and attire.
Addressing this critical challenge becomes imperative when harnessing gait for depression risk recognition.

\section{Conclusion}
\label{sec:Conclusion}

% 为了在初诊环境下给抑郁风险识别提供高效、易实现的抑郁风险识别方案，缓解抑郁症患者就诊率的不足，本文建立了迄今为止最大的基于步态的抑郁风险识别数据库。
% 同时还基于深度学习方法设计了一个以动态特征建模为核心的抑郁风险识别模型。
% 在所建立的大规模数据集上，通过实验验证了抑郁风险与步态之间的关联，进一步证明通过步态进行抑郁风险识别是可行的。
% 此外，本文通过实验对抑郁风险相关的步态特征进行了分析，验证了抑郁风险相关特征主要是体现在时间维度上的动态特征.
% 并且分析了不同视角对抑郁风险相关特征的影响，以及不同严重程度的抑郁风险所关联的步态特征。
To provide an efficient and implementable solution for recognizing depression risks in the primary care setting, this study establishes the largest gait-based depression risk identification database to date. 
Additionally, a depression risk recognition model focused on dynamic feature modeling is designed based on deep learning methodology. 
Through experiments conducted on the constructed large-scale dataset, the correlation between depression risk and gait is validated, further demonstrating the feasibility of utilizing gait for depression risk recognition.
This study analyzes the gait features related to depression risk and verifies that these features primarily manifest as dynamic characteristics in the temporal dimension.
Furthermore, an analysis is conducted to examine the impact of different perspectives and the severity grading of depression risk on the features related to depression risk.

%%

% 本文是对基于步态进行抑郁风险识别的一次初步尝试，为后续的研究者构建了数据基础，验证了这一任务的可行性。
% 并对抑郁风险相关的步态特征进行了初步探索，并给出了一些基本的结论
% 能够帮助后续的研究者们了解这一领域的特点，并在此基础上开展进一步的研究。
% 这一领域的每一次进步最终都将帮助正在经历痛苦的人们。
This paper represents an attempt to explore depression risk recognition through gait analysis, providing a foundational dataset for future researchers and validating the feasibility of this task.
Preliminary investigations into gait features associated with depression risk are conducted, and some basic conclusions are drawn.
These findings can assist subsequent researchers in understanding the characteristics of this field and conducting further studies.
Every advancement in this field ultimately contributes to aiding individuals who are undergoing distress.

% ------------- to be continued -------------- %

%\section*{Acknowledgment}

%The preferred spelling of the word ``acknowledgment'' in American English is 
%without an ``e'' after the ``g.'' Use the singular heading even if you have 
%many acknowledgments. Avoid expressions such as ``One of us (S.B.A.) would 
%like to thank $\ldots$ .'' Instead, write ``F. A. Author thanks $\ldots$ .'' In most 
%cases, sponsor and financial support acknowledgments are placed in the 
%unnumbered footnote on the first page, not here.

\section*{References}
\vspace{-0.6cm}
\bibliographystyle{IEEEtran}
\bibliography{main}

\begin{thebibliography}{10}
\providecommand{\url}[1]{#1}
\csname url@samestyle\endcsname
\providecommand{\newblock}{\relax}
\providecommand{\bibinfo}[2]{#2}
\providecommand{\BIBentrySTDinterwordspacing}{\spaceskip=0pt\relax}
\providecommand{\BIBentryALTinterwordstretchfactor}{4}
\providecommand{\BIBentryALTinterwordspacing}{\spaceskip=\fontdimen2\font plus
\BIBentryALTinterwordstretchfactor\fontdimen3\font minus
  \fontdimen4\font\relax}
\providecommand{\BIBforeignlanguage}[2]{{%
\expandafter\ifx\csname l@#1\endcsname\relax
\typeout{** WARNING: IEEEtran.bst: No hyphenation pattern has been}%
\typeout{** loaded for the language `#1'. Using the pattern for}%
\typeout{** the default language instead.}%
\else
\language=\csname l@#1\endcsname
\fi
#2}}
\providecommand{\BIBdecl}{\relax}
\BIBdecl

\bibitem{WHO}
W.~H. Organization, ``Mental health,''
  \url{https://www.who.int/health-topics/mental-health#tab=tab_1}, Mar. 2023.

\bibitem{vizhub}
I.~of~Health~Metrics and Evaluation, ``Global health data exchange (ghdx),''
  \url{https://vizhub.healthdata.org/gbd-results/}, Mar. 2023.

\bibitem{kennedy2001quality}
S.~H. Kennedy, B.~S. Eisfeld, and R.~G. Cooke, ``Quality of life: an important
  dimension in assessing the treatment of depression?'' \emph{Journal of
  psychiatry \& neuroscience: JPN}, vol.~26, no. Suppl, p. S23, 2001.

\bibitem{gaynes2002depression}
B.~N. Gaynes, B.~J. Burns, D.~L. Tweed, and P.~Erickson, ``Depression and
  health-related quality of life,'' \emph{The Journal of nervous and mental
  disease}, vol. 190, no.~12, pp. 799--806, 2002.

\bibitem{hirschfeld2000social}
R.~Hirschfeld, S.~A. Montgomery, M.~B. Keller, S.~Kasper, A.~F. Schatzberg,
  M.~Hans-Jurgen, D.~Healy, D.~Baldwin, M.~Humble, and M.~Versiani, ``Social
  functioning in depression: a review,'' \emph{Journal of Clinical Psychiatry},
  vol.~61, no.~4, pp. 268--275, 2000.

\bibitem{hansson2002quality}
L.~Hansson, ``Quality of life in depression and anxiety,'' \emph{International
  Review of Psychiatry}, vol.~14, no.~3, pp. 185--189, 2002.

\bibitem{bourassa2017social}
K.~J. Bourassa, M.~Memel, C.~Woolverton, and D.~A. Sbarra, ``Social
  participation predicts cognitive functioning in aging adults over time:
  comparisons with physical health, depression, and physical activity,''
  \emph{Aging \& mental health}, vol.~21, no.~2, pp. 133--146, 2017.

\bibitem{roshanaei2009longitudinal}
B.~Roshanaei-Moghaddam, W.~J. Katon, and J.~Russo, ``The longitudinal effects
  of depression on physical activity,'' \emph{General hospital psychiatry},
  vol.~31, no.~4, pp. 306--315, 2009.

\bibitem{hawton2013risk}
K.~Hawton, C.~C. i~Comabella, C.~Haw, and K.~Saunders, ``Risk factors for
  suicide in individuals with depression: a systematic review,'' \emph{Journal
  of affective disorders}, vol. 147, no. 1-3, pp. 17--28, 2013.

\bibitem{center2003confronting}
C.~Center, M.~Davis, T.~Detre, D.~E. Ford, W.~Hansbrough, H.~Hendin, J.~Laszlo,
  D.~A. Litts, J.~Mann, P.~A. Mansky \emph{et~al.}, ``Confronting depression
  and suicide in physicians: a consensus statement,'' \emph{Jama}, vol. 289,
  no.~23, pp. 3161--3166, 2003.

\bibitem{coulehan1997treating}
J.~L. Coulehan, H.~C. Schulberg, M.~R. Block, M.~J. Madonia, and E.~Rodriguez,
  ``Treating depressed primary care patients improves their physical, mental,
  and social functioning,'' \emph{Archives of Internal Medicine}, vol. 157,
  no.~10, pp. 1113--1120, 1997.

\bibitem{coyne1994prevalence}
J.~C. Coyne, S.~Fechner-Bates, and T.~L. Schwenk, ``Prevalence, nature, and
  comorbidity of depressive disorders in primary care,'' \emph{General hospital
  psychiatry}, vol.~16, no.~4, pp. 267--276, 1994.

\bibitem{halfin2007depression}
A.~Halfin, ``Depression: the benefits of early and appropriate treatment,''
  \emph{American Journal of Managed Care}, vol.~13, no.~4, p. S92, 2007.

\bibitem{weich2007attitudes}
S.~Weich, L.~Morgan, M.~King, and I.~Nazareth, ``Attitudes to depression and
  its treatment in primary care,'' \emph{Psychological medicine}, vol.~37,
  no.~9, pp. 1239--1248, 2007.

\bibitem{souery2007clinical}
D.~Souery, P.~Oswald, I.~Massat, U.~Bailer, J.~Bollen, K.~Demyttenaere,
  S.~Kasper, Y.~Lecrubier, S.~Montgomery, A.~Serretti \emph{et~al.}, ``Clinical
  factors associated with treatment resistance in major depressive disorder:
  results from a european multicenter study.'' \emph{Journal of Clinical
  Psychiatry}, vol.~68, no.~7, pp. 1062--1070, 2007.

\bibitem{akincigil2017national}
A.~Akincigil and E.~B. Matthews, ``National rates and patterns of depression
  screening in primary care: results from 2012 and 2013,'' \emph{Psychiatric
  services}, vol.~68, no.~7, pp. 660--666, 2017.

\bibitem{barbui2006identification}
C.~Barbui and M.~Tansella, ``Identification and management of depression in
  primary care settings. a meta-review of evidence,'' \emph{Epidemiology and
  Psychiatric Sciences}, vol.~15, no.~4, pp. 276--283, 2006.

\bibitem{young2001quality}
A.~S. Young, R.~Klap, C.~D. Sherbourne, and K.~B. Wells, ``The quality of care
  for depressive and anxiety disorders in the united states,'' \emph{Archives
  of general psychiatry}, vol.~58, no.~1, pp. 55--61, 2001.

\bibitem{kessler2005prevalence}
R.~C. Kessler, O.~Demler, R.~G. Frank, M.~Olfson, H.~A. Pincus, E.~E. Walters,
  P.~Wang, K.~B. Wells, and A.~M. Zaslavsky, ``Prevalence and treatment of
  mental disorders, 1990 to 2003,'' \emph{New England Journal of Medicine},
  vol. 352, no.~24, pp. 2515--2523, 2005.

\bibitem{belmaker2008major}
R.~H. Belmaker and G.~Agam, ``Major depressive disorder,'' \emph{New England
  Journal of Medicine}, vol. 358, no.~1, pp. 55--68, 2008.

\bibitem{greden2001burden}
J.~F. Greden, ``The burden of recurrent depression: causes, consequences, and
  future prospects,'' \emph{Journal of Clinical Psychiatry}, vol.~62, pp. 5--9,
  2001.

\bibitem{mohr2006barriers}
D.~C. Mohr, S.~L. Hart, I.~Howard, L.~Julian, L.~Vella, C.~Catledge, and M.~D.
  Feldman, ``Barriers to psychotherapy among depressed and nondepressed primary
  care patients,'' \emph{Annals of Behavioral Medicine}, vol.~32, no.~3, pp.
  254--258, 2006.

\bibitem{kravitz2011relational}
R.~L. Kravitz, D.~A. Paterniti, R.~M. Epstein, A.~B. Rochlen, R.~A. Bell,
  C.~Cipri, E.~F. y~Garcia, M.~D. Feldman, and P.~Duberstein, ``Relational
  barriers to depression help-seeking in primary care,'' \emph{Patient
  education and counseling}, vol.~82, no.~2, pp. 207--213, 2011.

\bibitem{cassano2002depression}
P.~Cassano and M.~Fava, ``Depression and public health: an overview,''
  \emph{Journal of psychosomatic research}, vol.~53, no.~4, pp. 849--857, 2002.

\bibitem{bell2011suffering}
R.~A. Bell, P.~Franks, P.~R. Duberstein, R.~M. Epstein, M.~D. Feldman, E.~F.
  y~Garcia, and R.~L. Kravitz, ``Suffering in silence: reasons for not
  disclosing depression in primary care,'' \emph{The Annals of Family
  Medicine}, vol.~9, no.~5, pp. 439--446, 2011.

\bibitem{johnstone2001stigma}
M.-J. Johnstone, ``Stigma, social justice and the rights of the mentally ill:
  Challenging the status quo,'' \emph{Australian and New Zealand Journal of
  Mental Health Nursing}, vol.~10, no.~4, pp. 200--209, 2001.

\bibitem{moyle2002unstructured}
W.~Moyle, ``Unstructured interviews: challenges when participants have a major
  depressive illness,'' \emph{Journal of advanced nursing}, vol.~39, no.~3, pp.
  266--273, 2002.

\bibitem{delbaere2010determinants}
K.~Delbaere, J.~C. Close, H.~Brodaty, P.~Sachdev, and S.~R. Lord,
  ``Determinants of disparities between perceived and physiological risk of
  falling among elderly people: cohort study,'' \emph{Bmj}, vol. 341, 2010.

\bibitem{rost2004effect}
K.~Rost, J.~L. Smith, and M.~Dickinson, ``The effect of improving primary care
  depression management on employee absenteeism and productivity a randomized
  trial,'' \emph{Medical care}, vol.~42, no.~12, p. 1202, 2004.

\bibitem{docherty1997barriers}
J.~P. Docherty, ``Barriers to the diagnosis of depression in primary care,''
  \emph{Journal of clinical psychiatry}, vol.~58, no.~1, pp. 5--10, 1997.

\bibitem{dowrick2013medicalising}
C.~Dowrick and A.~Frances, ``Medicalising unhappiness: new classification of
  depression risks more patients being put on drug treatment from which they
  will not benefit,'' \emph{bmj}, vol. 347, 2013.

\bibitem{cheng2021addressing}
N.~Cheng and S.~Mohiuddin, ``Addressing the nationwide shortage of child and
  adolescent psychiatrists: determining factors that influence the decision for
  psychiatry residents to pursue child and adolescent psychiatry training,''
  \emph{Academic psychiatry}, pp. 1--7, 2021.

\bibitem{butryn2017shortage}
T.~Butryn, L.~Bryant, C.~Marchionni, and F.~Sholevar, ``The shortage of
  psychiatrists and other mental health providers: causes, current state, and
  potential solutions,'' \emph{International Journal of Academic Medicine},
  vol.~3, no.~1, pp. 5--9, 2017.

\bibitem{kanai2003time}
T.~Kanai, H.~Takeuchi, T.~A. Furukawa, R.~Yoshimura, T.~Imaizumi, T.~Kitamura,
  and K.~Takahashi, ``Time to recurrence after recovery from major depressive
  episodes and its predictors,'' \emph{Psychological Medicine}, vol.~33, no.~5,
  pp. 839--845, 2003.

\bibitem{sobin1997psychomotor}
C.~Sobin and H.~A. Sackeim, ``Psychomotor symptoms of depression,''
  \emph{American Journal of Psychiatry}, vol. 154, no.~1, pp. 4--17, 1997.

\bibitem{schrijvers2008psychomotor}
D.~Schrijvers, W.~Hulstijn, and B.~G. Sabbe, ``Psychomotor symptoms in
  depression: a diagnostic, pathophysiological and therapeutic tool,''
  \emph{Journal of affective disorders}, vol. 109, no. 1-2, pp. 1--20, 2008.

\bibitem{fried2014impact}
E.~I. Fried and R.~M. Nesse, ``The impact of individual depressive symptoms on
  impairment of psychosocial functioning,'' \emph{PloS one}, vol.~9, no.~2, p.
  e90311, 2014.

\bibitem{takakusaki2017functional}
K.~Takakusaki, ``Functional neuroanatomy for posture and gait control,''
  \emph{Journal of movement disorders}, vol.~10, no.~1, p.~1, 2017.

\bibitem{walther2019utility}
S.~Walther, J.~Bernard, V.~Mittal, and S.~Shankman, ``The utility of an rdoc
  motor domain to understand psychomotor symptoms in depression,''
  \emph{Psychological medicine}, vol.~49, no.~2, pp. 212--216, 2019.

\bibitem{michalak2009embodiment}
J.~Michalak, N.~F. Troje, J.~Fischer, P.~Vollmar, T.~Heidenreich, and
  D.~Schulte, ``Embodiment of sadness and depression—gait patterns associated
  with dysphoric mood,'' \emph{Psychosomatic medicine}, vol.~71, no.~5, pp.
  580--587, 2009.

\bibitem{lemke2000spatiotemporal}
M.~R. Lemke, T.~Wendorff, B.~Mieth, K.~Buhl, and M.~Linnemann, ``Spatiotemporal
  gait patterns during over ground locomotion in major depression compared with
  healthy controls,'' \emph{Journal of psychiatric research}, vol.~34, no. 4-5,
  pp. 277--283, 2000.

\bibitem{nilsonne2021eeg}
G.~Nilsonne and F.~E. Harrell~Jr, ``Eeg-based model and antidepressant
  response,'' \emph{Nature Biotechnology}, vol.~39, no.~1, pp. 27--27, 2021.

\bibitem{schnyer2017evaluating}
D.~M. Schnyer, P.~C. Clasen, C.~Gonzalez, and C.~G. Beevers, ``Evaluating the
  diagnostic utility of applying a machine learning algorithm to diffusion
  tensor mri measures in individuals with major depressive disorder,''
  \emph{Psychiatry Research: Neuroimaging}, vol. 264, pp. 1--9, 2017.

\bibitem{ramasubbu2016accuracy}
R.~Ramasubbu, M.~R. Brown, F.~Cortese, I.~Gaxiola, B.~Goodyear, A.~J.
  Greenshaw, S.~M. Dursun, and R.~Greiner, ``Accuracy of automated
  classification of major depressive disorder as a function of symptom
  severity,'' \emph{NeuroImage: Clinical}, vol.~12, pp. 320--331, 2016.

\bibitem{vai2020predicting}
B.~Vai, L.~Parenti, I.~Bollettini, C.~Cara, C.~Verga, E.~Melloni, E.~Mazza,
  S.~Poletti, C.~Colombo, and F.~Benedetti, ``Predicting differential diagnosis
  between bipolar and unipolar depression with multiple kernel learning on
  multimodal structural neuroimaging,'' \emph{European
  Neuropsychopharmacology}, vol.~34, pp. 28--38, 2020.

\bibitem{shim2019machine}
M.~Shim, M.~J. Jin, C.-H. Im, and S.-H. Lee, ``Machine-learning-based
  classification between post-traumatic stress disorder and major depressive
  disorder using p300 features,'' \emph{NeuroImage: Clinical}, vol.~24, p.
  102001, 2019.

\bibitem{jiang2016predictability}
H.~Jiang, T.~Popov, P.~Jyl{\"a}nki, K.~Bi, Z.~Yao, Q.~Lu, O.~Jensen, and
  M.~Van~Gerven, ``Predictability of depression severity based on posterior
  alpha oscillations,'' \emph{Clinical Neurophysiology}, vol. 127, no.~4, pp.
  2108--2114, 2016.

\bibitem{sato2015machine}
J.~R. Sato, J.~Moll, S.~Green, J.~F. Deakin, C.~E. Thomaz, and R.~Zahn,
  ``Machine learning algorithm accurately detects fmri signature of
  vulnerability to major depression,'' \emph{Psychiatry Research:
  Neuroimaging}, vol. 233, no.~2, pp. 289--291, 2015.

\bibitem{wei2021functional}
Y.~Wei, Q.~Chen, A.~Curtin, L.~Tu, X.~Tang, Y.~Tang, L.~Xu, Z.~Qian, J.~Zhou,
  C.~Zhu \emph{et~al.}, ``Functional near-infrared spectroscopy (fnirs) as a
  tool to assist the diagnosis of major psychiatric disorders in a chinese
  population,'' \emph{European archives of psychiatry and clinical
  neuroscience}, vol. 271, pp. 745--757, 2021.

\bibitem{zhou2018visually}
X.~Zhou, K.~Jin, Y.~Shang, and G.~Guo, ``Visually interpretable representation
  learning for depression recognition from facial images,'' \emph{IEEE
  transactions on affective computing}, vol.~11, no.~3, pp. 542--552, 2018.

\bibitem{ma2016depaudionet}
X.~Ma, H.~Yang, Q.~Chen, D.~Huang, and Y.~Wang, ``Depaudionet: An efficient
  deep model for audio based depression classification,'' in \emph{Proceedings
  of the 6th international workshop on audio/visual emotion challenge}, 2016,
  pp. 35--42.

\bibitem{wang2004fusion}
L.~Wang, H.~Ning, T.~Tan, and W.~Hu, ``Fusion of static and dynamic body
  biometrics for gait recognition,'' \emph{IEEE Transactions on circuits and
  systems for video technology}, vol.~14, no.~2, pp. 149--158, 2004.

\bibitem{li2020end}
X.~Li, Y.~Makihara, C.~Xu, Y.~Yagi, S.~Yu, and M.~Ren, ``End-to-end model-based
  gait recognition,'' in \emph{Proceedings of the Asian conference on computer
  vision}, 2020.

\bibitem{kusakunniran2010support}
W.~Kusakunniran, Q.~Wu, J.~Zhang, and H.~Li, ``Support vector regression for
  multi-view gait recognition based on local motion feature selection,'' in
  \emph{2010 IEEE Computer society conference on computer vision and pattern
  recognition}.\hskip 1em plus 0.5em minus 0.4em\relax IEEE, 2010, pp.
  974--981.

\bibitem{bashir2010cross}
K.~Bashir, T.~Xiang, and S.~Gong, ``Cross view gait recognition using
  correlation strength.'' in \emph{Bmvc}, 2010, pp. 1--11.

\bibitem{wu2016comprehensive}
Z.~Wu, Y.~Huang, L.~Wang, X.~Wang, and T.~Tan, ``A comprehensive study on
  cross-view gait based human identification with deep cnns,'' \emph{IEEE
  transactions on pattern analysis and machine intelligence}, vol.~39, no.~2,
  pp. 209--226, 2016.

\bibitem{yu2017gaitgan}
S.~Yu, H.~Chen, E.~B. Garcia~Reyes, and N.~Poh, ``Gaitgan: Invariant gait
  feature extraction using generative adversarial networks,'' in
  \emph{Proceedings of the IEEE conference on computer vision and pattern
  recognition workshops}, 2017, pp. 30--37.

\bibitem{chao2019gaitset}
H.~Chao, Y.~He, J.~Zhang, and J.~Feng, ``Gaitset: Regarding gait as a set for
  cross-view gait recognition,'' in \emph{Proceedings of the AAAI conference on
  artificial intelligence}, vol.~33, no.~01, 2019, pp. 8126--8133.

\bibitem{song2019gaitnet}
C.~Song, Y.~Huang, Y.~Huang, N.~Jia, and L.~Wang, ``Gaitnet: An end-to-end
  network for gait based human identification,'' \emph{Pattern recognition},
  vol.~96, p. 106988, 2019.

\bibitem{fan2020gaitpart}
C.~Fan, Y.~Peng, C.~Cao, X.~Liu, S.~Hou, J.~Chi, Y.~Huang, Q.~Li, and Z.~He,
  ``Gaitpart: Temporal part-based model for gait recognition,'' in
  \emph{Proceedings of the IEEE/CVF conference on computer vision and pattern
  recognition}, 2020, pp. 14\,225--14\,233.

\bibitem{hou2020gait}
S.~Hou, C.~Cao, X.~Liu, and Y.~Huang, ``Gait lateral network: Learning
  discriminative and compact representations for gait recognition,'' in
  \emph{European conference on computer vision}.\hskip 1em plus 0.5em minus
  0.4em\relax Springer, 2020, pp. 382--398.

\bibitem{lin2021gait}
B.~Lin, S.~Zhang, and X.~Yu, ``Gait recognition via effective global-local
  feature representation and local temporal aggregation,'' in \emph{Proceedings
  of the IEEE/CVF International Conference on Computer Vision}, 2021, pp.
  14\,648--14\,656.

\bibitem{sloman1982gait}
L.~Sloman, M.~Berridge, S.~Homatidis, D.~Hunter, and T.~Duck, ``Gait patterns
  of depressed patients and normal subjects.'' \emph{The American journal of
  psychiatry}, vol. 139, no.~1, pp. 94--97, 1982.

\bibitem{wang2020gait}
T.~Wang, C.~Li, C.~Wu, C.~Zhao, J.~Sun, H.~Peng, X.~Hu, and B.~Hu, ``A gait
  assessment framework for depression detection using kinect sensors,''
  \emph{IEEE Sensors Journal}, vol.~21, no.~3, pp. 3260--3270, 2020.

\bibitem{lu2021new}
H.~Lu, W.~Shao, E.~Ngai, X.~Hu, and B.~Hu, ``A new skeletal representation
  based on gait for depression detection,'' in \emph{2020 IEEE International
  Conference on E-health Networking, Application \& Services
  (HEALTHCOM)}.\hskip 1em plus 0.5em minus 0.4em\relax IEEE, 2021, pp. 1--6.

\bibitem{fang2019depression}
J.~Fang, T.~Wang, C.~Li, X.~Hu, E.~Ngai, B.-C. Seet, J.~Cheng, Y.~Guo, and
  X.~Jiang, ``Depression prevalence in postgraduate students and its
  association with gait abnormality,'' \emph{IEEE Access}, vol.~7, pp.
  174\,425--174\,437, 2019.

\bibitem{yuan2019depression}
Y.~Yuan, B.~Li, N.~Wang, Q.~Ye, Y.~Liu, and T.~Zhu, ``Depression identification
  from gait spectrum features based on hilbert-huang transform,'' in
  \emph{Human Centered Computing: 4th International Conference, HCC 2018,
  M{\'e}rida, Mexico, December, 5--7, 2018, Revised Selected Papers 4}.\hskip
  1em plus 0.5em minus 0.4em\relax Springer, 2019, pp. 503--515.

\end{thebibliography}

\end{document}